\documentclass[11pt]{article}

\usepackage[preprint]{acl}

\usepackage{times}
\usepackage{latexsym}

\usepackage[T1]{fontenc}

\usepackage[utf8]{inputenc}

\usepackage{microtype}

\usepackage{inconsolata}

\usepackage{graphicx}
\usepackage{array}

\usepackage{tikz}
\usetikzlibrary{arrows.meta,backgrounds}
\usepackage{pgfplots}
\pgfplotsset{compat=1.18}
\definecolor{vbaccent}{HTML}{FF3C00}%
\definecolor{vbink}{HTML}{181818}%
\definecolor{vbdark}{HTML}{312E2E}%
\definecolor{vbmuted}{HTML}{767270}%
\definecolor{vbsilver}{HTML}{C5C5C5}%
\definecolor{vbgrid}{HTML}{E2E2E2}%
\colorlet{vbfill}{vbgrid}%
\definecolor{vbpale}{HTML}{FFB199}%
\definecolor{vbgreen}{HTML}{1E9E55}%

\newcommand{\platform}{Replit}

\title{ViBench 2.0: A Benchmark for Long-Horizon Software Building}

\author{
  \textbf{Daniel Furman\textsuperscript{*}},
  \textbf{Preeya Kirani\textsuperscript{*}},
  \textbf{Ibrahim Sheikh},
  \textbf{Zhen Li},
  \textbf{Peter Zhong},
  \textbf{Michele Catasta}
\\
  Replit
\\
  \small{\textsuperscript{*}Equal contribution, listed alphabetically. \textbf{Correspondence:} \href{mailto:daniel.furman@repl.it}{daniel.furman@repl.it}}
}

\begin{document}
\maketitle
\begin{abstract}
On \platform{}, our agentic engineering platform, users now build entire products without writing code, through successive feature requests to an agent.
Yet existing coding benchmarks evaluate single, fully specified tasks, and many are near saturation.
We introduce ViBench 2.0, a benchmark for long-horizon software building that we use to select our platform's models.
Its 17 full-stack web products, modeled on widely used software, grow over 241 feature requests and are judged only when finished, as a user would judge them: a browser agent runs 544 test plans, and a build passes only if every plan does, under a grading protocol validated to keep this strict score reliable.
Evaluating nine models, we find a reliability gap: the three strongest pass up to 98\% of test plans yet deliver a fully working product in at most 65\% of builds, and no open-weight model delivers one.
The gap is closing fast: across three releases of one model family, pass@1 on the public products rises from 3\% to 53\% while the cost per build falls by more than half.
We release eight products with the harness and grader, hold out nine, and report lessons from production use.\footnote{Code and data: \url{https://github.com/ViBench/vibench-public}.}
\end{abstract}

\section{Introduction}
\label{sec:intro}

Consider a ride-hailing app that an agent builds from an empty project: riders first book rides, and later requests add cancellation fees, drivers with several cars, per-trip earnings and split fares, each built on code the agent wrote earlier 
(Figure~\ref{fig:deps}).
Projects on \platform{} start greenfield, from an empty project, and grow this way through agentic engineering, in which a user describes software in natural language and a coding agent builds and extends it, one request at a time; users who work this way favor iteration over written specifications~\citep{pimenova2025goodvibes}.
Agentic engineering is what vibe coding~\citep{karpathy2025vibe}, from which ViBench~\citep{zhong2026vibench} takes its name, has matured into: agents that once delivered near-functional prototypes now build, extend and maintain the products businesses run.
Agents can now sustain such long-horizon work for hours or days~\citep{kwa2025timehorizon,carlini2026compiler,lin2026scaling}, yet the benchmarks used to compare them saturate quickly: over 40\% of those less than two years old have already saturated~\citep{akhtar2026plateau}.

Agentic engineering benchmarks are no exception.
ViBench~\citep{zhong2026vibench} has an agent build a minimum viable product (MVP) or add one feature to it, and a browser agent test the result through its interface; its pass@1~\citep{chen2021codex} counts a build only if every test plan passes.
In our rerun of its Vibe-on-Vibe tasks, in which the agent extends its own MVP, GPT-6.1 Sol, GPT-6 Astra and Opus 5.5 (hereafter Sol, Astra and Opus) reach 82 to 89\% pass@1, against 51\% for the best model at its release (Appendix~\ref{app:rerun}).
On Vibe Code Bench, in which an agent builds an app from one specification, the best score rose from 61.8\% in March 2026~\citep{tran2026vibecodebench} to over 90\% for six models by late October~\citep{vals2026vibecodebench}.

Like most coding benchmarks~\citep{jimenez2024swebench,deng2025swebenchpro,zhao2025commit0,ding2026nl2repobench}, both give the agent a single task, while multi-turn benchmarks score each step as it is completed~\citep{xu2025webbench,orlanski2026slopcodebench,shen2026evocodebench}.
Vibe Code Bench 1-100~\citep{tran2026vcb1100}, released in September 2026, grows an application over up to ten requests and tops out at 34.0\%~\citep{vals2026vibecodebench}, but from a finished reference app and only until the first failure, so it measures how far an agent gets rather than what it delivers.
None of these asks the question such users face: whether a product built from nothing still works as a whole after the last request (Appendix~\ref{app:related}).
Such a product can only be checked through its interface, the way users who never read the code judge it~\citep{karpathy2025vibe,fawzy2025vibepractice,sarkar2025vibe}, and a build should count only if every plan passes.

We introduce ViBench 2.0, which extends ViBench to evaluate whether agents can build a complete product over a long horizon.
The agent receives a product's requirements one feature at a time in a single conversation, starting from an empty project, and a browser agent tests only the finished product (Figure~\ref{fig:pipeline}).
The horizon is long in three ways: each of 9 to 18 feature requests builds on code the agent wrote itself, the conversation can exceed 200k tokens, and every feature must survive every request after it, with no feedback between requests.
Across 17 products, ViBench 2.0 comprises 241 feature requests and 544 test plans, 85\% of which combine features. ViBench 2.0 is built for deployment decisions: we use it to choose the models \platform{} serves and to evaluate changes to its production agent.

This paper advances the evaluation of coding agents in four ways.
First, a benchmark that measures frontier models on greenfield software building without saturating: pass@1 spans 0 to 65\% across nine models and rises across successive versions of one model family; we release eight products on Harbor~\citep{harbor2026} and hold out nine.
Second, a grading protocol, extending the grader of ViBench~\citep{zhong2026vibench}, that makes an all-or-nothing score reliable: it re-grades failed plans until three of up to five grades agree, caught all 34 planted bugs, and only 0.8\% of its passes are false (Section~\ref{sec:validation}).
Third, independent runs of all nine models: an external team at Fireworks~\citep{fireworks2026sii} set up the released benchmark and ran each model with one command, with no other changes. These runs show that the headroom lies in reliability: most failed builds of the top three models miss only one or two plans (Section~\ref{sec:results}).
Fourth, lessons from using the benchmark across \platform{}'s deployment decisions, from which models serve its workflows to how its agent manages context (Section~\ref{sec:practice}).

\section{Benchmark Design}
\label{sec:benchmark}

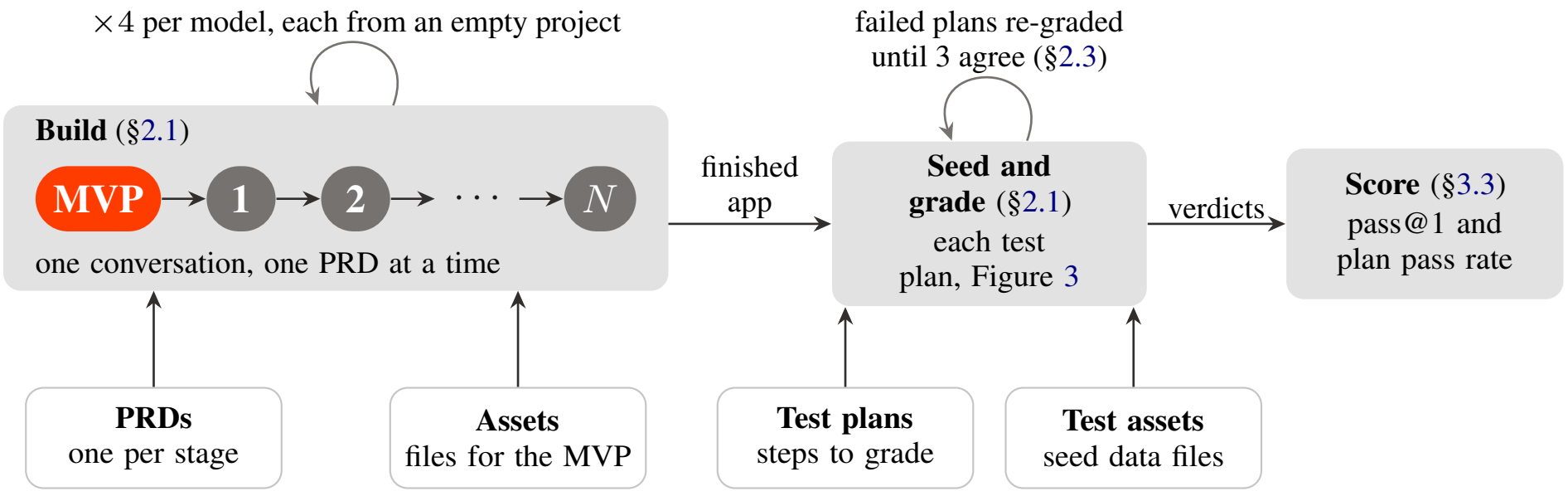
\begin{figure*}[t]
  \centering
  \begin{tikzpicture}[
    x=1cm, y=1cm,
    pbox/.style={fill=vbfill, rounded corners=4pt, font=\scriptsize, align=center, inner sep=3pt,
      text width=#1, minimum height=1.2cm},
    pbox/.default=2.3cm,
    pin/.style={draw=vbsilver, fill=white, rounded corners=4pt, font=\scriptsize, align=center, inner sep=2pt,
      text width=1.9cm, minimum height=0.8cm},
    pstage/.style={fill=vbmuted, draw=none, text=white, rounded corners=2.6mm, font=\footnotesize\bfseries, inner xsep=4pt, inner ysep=3pt,
      minimum height=0.52cm, minimum width=0.55cm, text depth=0pt},
    pflow/.style={-{Stealth[length=4pt, width=3.5pt]}, line width=0.5pt, draw=vbdark},
    ploop/.style={pflow, draw=vbmuted},
    pnote/.style={font=\scriptsize, align=center, inner sep=1pt},
  ]
  \coordinate (origin) at (0, 0);
  \coordinate (inrow) at (0, -1.3);

  \node[pstage, anchor=west, minimum width=0.8cm, fill=vbaccent] (s0) at ([xshift=2.5mm, yshift=2mm]origin) {MVP};
  \node[pstage, anchor=west] (s1) at ([xshift=3.5mm]s0.east) {1};
  \node[pstage, anchor=west] (s2) at ([xshift=3.5mm]s1.east) {2};
  \node[pstage, anchor=west, fill=none, text=vbink, minimum width=0.45cm] (sdots) at ([xshift=3.5mm]s2.east) {$\cdots$};
  \node[pstage, anchor=west] (sN) at ([xshift=3.5mm]sdots.east) {$N$};
  \foreach \a/\b in {s0/s1, s1/s2, s2/sdots, sdots/sN}
    \draw[pflow] (\a) -- (\b);
  \node[pnote, anchor=south west, inner sep=0pt] (btitle) at ([yshift=1.5mm]s0.north west)
    {\textbf{Build} (\S\ref{sec:environment})};
  \node[pnote, anchor=north west, inner sep=0pt, align=left, text width=4.3cm] (bnote) at ([yshift=-1.5mm]s0.south west)
    {one conversation, one PRD at a time};
  \coordinate (s0w) at ([xshift=-2.5mm]s0.west);
  \coordinate (sNe) at ([xshift=2.5mm]sN.east);
  \coordinate (bl) at (s0w |- origin);
  \coordinate (br) at (sNe |- origin);
  \coordinate (bt) at ([yshift=1mm]btitle.north);
  \coordinate (bb) at ([yshift=-1mm]bnote.south);
  \begin{scope}[on background layer]
    \fill[vbfill, rounded corners=4pt] (bl |- bt) rectangle (br |- bb);
  \end{scope}
  \coordinate (bloop) at (s2.north |- bt);
  \draw[ploop] ([xshift=3mm]bloop) to[out=70, in=110, looseness=3]
    node[pnote, above] {$\times 4$ per model, each from an empty project} ([xshift=-3mm]bloop);

  \node[pbox, anchor=west] (sg) at ([xshift=13mm]br)
    {\textbf{Seed and grade} (\S\ref{sec:seeding-grading})\\[1pt] each test plan, Figure~\ref{fig:seed-grade}};
  \draw[ploop] ([xshift=3mm]sg.north) to[out=70, in=110, looseness=3]
    node[pnote, above] {failed plans re-graded\\until 3 agree (\S\ref{sec:protocol})} ([xshift=-3mm]sg.north);
  \node[pbox=2.0cm, anchor=west] (score) at ([xshift=11mm]sg.east)
    {\textbf{Score} (\S\ref{sec:metrics})\\[1pt] pass@1 and plan pass rate};
  \draw[pflow] (br) -- node[pnote, above] {finished\\app} (sg.west);
  \draw[pflow] (sg) -- node[pnote, above] {verdicts} (score);

  \coordinate (prdsx) at ([xshift=12mm]bl);
  \coordinate (assetsx) at ([xshift=-12mm]br);
  \coordinate (plansx) at ([xshift=-11.5mm]sg.south);
  \coordinate (tassetsx) at ([xshift=11.5mm]sg.south);
  \node[pin, anchor=north] (prds) at (prdsx |- inrow) {\textbf{PRDs}\\ one per stage};
  \node[pin, anchor=north] (assets) at (assetsx |- inrow) {\textbf{Assets}\\ files for the MVP};
  \node[pin, anchor=north] (plans) at (plansx |- inrow) {\textbf{Test plans}\\ steps to grade};
  \node[pin, anchor=north] (tassets) at (tassetsx |- inrow) {\textbf{Test assets}\\ seed data files};
  \draw[pflow] (prds.north) -- (prds.north |- bb);
  \draw[pflow] (assets.north) -- (assets.north |- bb);
  \draw[pflow] (plans.north) -- (plans.north |- sg.south);
  \draw[pflow] (tassets.north) -- (tassets.north |- sg.south);
\end{tikzpicture}
  \caption{The ViBench 2.0 pipeline.
    The builder receives one PRD per stage and never sees the test plans; a seeding agent then loads each plan's data, and a browser agent grades the finished app.}
  \label{fig:pipeline}
\end{figure*}

An agent grows each product one request at a time, and only the finished product is graded, as a user would use it (Figure~\ref{fig:pipeline}).
We first define the task (Section~\ref{sec:environment}), then describe how we chose the products and wrote their PRDs and test plans (Section~\ref{sec:construction}), how we verified them (Section~\ref{sec:validation}) and what the 17 products contain (Section~\ref{sec:composition}).

\subsection{Task Formulation}
\label{sec:environment}

\paragraph{Building.}
The agent under evaluation, the builder, starts from an empty project and first receives the PRD for a minimum viable product (MVP; Figure~\ref{fig:pipeline}).
After each stage, it receives the next feature's PRD in the same conversation.
Each stage is limited to 300 agent steps and 2 hours, and beyond 200k tokens the agent's own model summarizes the older part of the conversation.
The agent works in a container with a PostgreSQL database and maintains two scripts: one that sets up the application and one that starts it.
ViBench 2.0 keeps the harness of ViBench~\citep{zhong2026vibench}, built on the OpenHands Software Agent SDK~\citep{wang2026openhands}, but runs it on Harbor~\citep{harbor2026}, the open evaluation framework behind Terminal-Bench 2.0~\citep{merrill2026terminalbench}.
Builds, seeds and grades run in fresh containers in parallel, so a full round is graded in five to six hours.
One command per model runs the whole pipeline, so other teams can run the benchmark unchanged, as the external team did (Appendix~\ref{app:setup}).

\paragraph{Evaluation.}
No grading takes place until the last stage is complete.
Because the agent chooses the architecture, the framework and the interface, no tests can be written in advance: unit tests presuppose a code structure, and scripted browser tests a page layout.
Evaluation therefore occurs at the level of an end user: test plans check only what a user of the finished product would see.
Plans grade outcomes rather than a specific design: wording, layout and the placement of controls are left to the agent, so a ride shown as ``Waiting at stop'' satisfies a PRD that calls the state ``At Stop''.

\paragraph{Seeding and grading.}
\label{sec:seeding-grading}
Every seed and grade starts a fresh container from the final code with an empty database, so a product must install and start cleanly, not merely function during development.
For each plan that requires starting data, a seeding agent inspects the finished application and writes a script that creates that data.
The grader then operates a real Chromium browser through Playwright~\citep{microsoft2020playwright} the way a user would, acting as a computer-use agent at screen coordinates when it cannot locate an element, and it writes its actions as it proceeds rather than replaying a fixed script, which would break on a build with a different interface (Appendix~\ref{app:grading}).
It judges each step solely from what the page presents to a user, cites that evidence, and never circumvents a defect; a plan passes only if every step passes.
For all builders, both the seeding agent and the grader use Opus 5.5 at medium effort.

\subsection{Task Construction}
\label{sec:construction}

\paragraph{Products.}
Each task is modeled on real software, and each stage adds a feature that the software ships, as described in its public help pages; the eight public products follow Amazon, Asana, Discord, Figma, GitHub, Google Docs, Jira and Uber.\footnote{Trademarks of their owners; the authors are not affiliated with these companies.}
Real products impose requirements beyond the primary path through an application: a chat message must reach everyone viewing the channel without a reload (Discord), a split fare must divide among the riders to the cent (Uber), and when two members move the same card from boards opened earlier, only the first move may take effect (Jira).

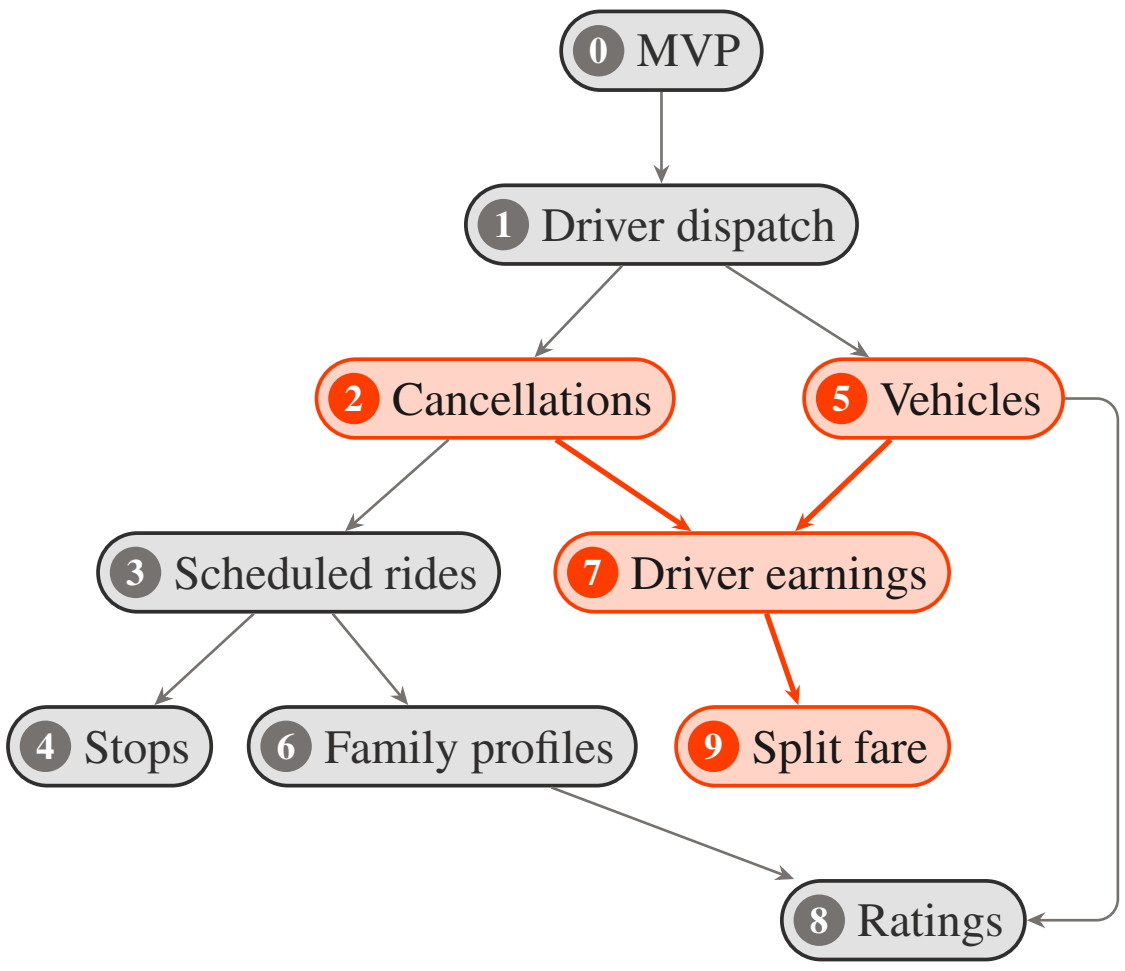
\begin{figure}[t]
  \centering
  \begin{tikzpicture}[
    x=1cm, y=1cm,
    stage/.style={fill=vbfill, draw=vbdark, line width=0.7pt, text=vbdark, rounded corners=2.6mm, font=\footnotesize,
      inner xsep=4pt, inner ysep=3pt, minimum height=0.52cm, text depth=0pt},
    hotstage/.style={stage, fill=vbpale!55, draw=vbaccent, line width=0.7pt, text=vbink},
    num/.style={circle, inner sep=0pt, minimum size=3.4mm, fill=vbmuted, text=white, font=\scriptsize\bfseries},
    hotnum/.style={num, fill=vbaccent, draw=none, text=white, font=\scriptsize\bfseries},
    dep/.style={-{Stealth[length=3.5pt, width=3pt]}, line width=0.5pt, draw=vbmuted},
    hot/.style={dep, draw=vbaccent, line width=1pt},
  ]
  \newcommand{\vbstage}[6][stage]{%
    \node[#1, anchor=center] (#2) at (#3) {\hspace{3.6mm}#5};
    \node[#6] at ([xshift=2.6mm]#2.west) {#4};}
  \vbstage{mvp}{4.40, 0.00}{0}{MVP}{num}
  \vbstage{dispatch}{4.40,-1.15}{1}{Driver dispatch}{num}
  \vbstage[hotstage]{cancel}{3.30,-2.30}{2}{Cancellations}{hotnum}
  \vbstage[hotstage]{vehicles}{6.20,-2.30}{5}{Vehicles}{hotnum}
  \vbstage{scheduled}{2.00,-3.45}{3}{Scheduled rides}{num}
  \vbstage[hotstage]{earnings}{5.00,-3.45}{7}{Driver earnings}{hotnum}
  \vbstage{stops}{0.75,-4.60}{4}{Stops}{num}
  \vbstage{family}{2.95,-4.60}{6}{Family profiles}{num}
  \vbstage[hotstage]{split}{5.40,-4.60}{9}{Split fare}{hotnum}
  \vbstage{ratings}{6.00,-5.75}{8}{Ratings}{num}

  \draw[dep] (mvp) -- (dispatch);
  \draw[dep] (dispatch) -- (cancel);
  \draw[dep] (cancel) -- (scheduled);
  \draw[dep] (scheduled) -- (family);
  \draw[dep] (family) -- (ratings);

  \draw[dep] (dispatch) -- (vehicles);
  \draw[hot] (cancel) -- (earnings);
  \draw[hot] (vehicles) -- (earnings);
  \draw[dep] (scheduled) -- (stops);
  \draw[hot] (earnings) -- (split);
  \draw[dep, rounded corners=4pt] (vehicles.east) -- ++(0.35, 0) |- (ratings.east);
\end{tikzpicture}
  \caption{Feature dependencies of the product modeled on Uber, numbered in build order; an arrow from A to B means that B builds on A.
    Orange marks features two plans check together: earnings must show the car and cancellation fees, and a split fare must not alter them.}
  \label{fig:deps}
\end{figure}

\paragraph{PRDs.}
Each PRD specifies only business-facing content, namely user flows, roles, business rules and the states a user can observe, and omits every technical detail, such as data models, APIs, schemas and libraries.
The first PRD describes the MVP; each subsequent PRD adds one feature that the real software ships, written as a change to what the agent has already built.
We ordered each product's features along a dependency graph, such that every feature arrives after those it builds on, and all models receive the same order (Figure~\ref{fig:deps}).

\paragraph{Test plans.}
A test plan is a sequence of steps that the grader executes as a user would, each step bundling a few related actions with the checks that follow them.
A few plans cover signing in or a single feature, but most exercise several at once: a late feature interacting with an early one, a page left open while another user makes a change, or two users acting at the same moment.
To prevent a single bug from costing more than one step, each behavior is graded by exactly one plan; other plans that encounter it merely record what happened.

\subsection{Verification}
\label{sec:validation}

We consider a task verified if it satisfies four criteria.

\textbf{Specificity.}
A plan passes if and only if the product behaves as its PRDs require.
Every check traces to an explicit statement in the PRD, a direct implication of one, or one of five basics that every agent's system prompt states and the PRDs therefore omit: the app starts on an empty database, nothing the user saves is lost, nothing shows as saved if it was not saved, one click does one thing, and a refused action shows a message.

\textbf{Solvability.}
It must be possible to build the product such that every plan passes.
Every product has been built to pass all of its plans.\footnote{At least one build of each product passes every plan, in the independent runs (Section~\ref{sec:results}) or in our reference runs.}

\textbf{Integrity.}
An agent must not be able to succeed through shortcuts.
Each PRD enters the builder's container only at its own stage, so no later feature is on disk, and the system prompt says only that more features will follow.
The builder never sees the test plans and has no web-search or browser tool, and the grader must judge the application as delivered, without repairing or circumventing defects.
Public files carry a canary string for filtering from training data, which the harness checks before every build.

\textbf{Grader accuracy.}
Because an agent performs the grading, its verdicts must be correct, and under an all-or-nothing score the two kinds of error, wrong passes and wrong fails, carry different costs: all 26 to 41 plans of a product must succeed, so one wrong fail costs a working build its pass.

\label{sec:protocol}
The grading protocol therefore targets wrong fails: a plan that passes its first grade is not graded again, but a plan that fails is re-graded up to four more times, until three grades agree, and a grade that lacks on-screen evidence or breaks a rule is repeated (Figure~\ref{fig:seed-grade}, Appendix~\ref{app:grading}).
We validated grading once, before release, on our own reference runs (Table~\ref{tab:grader}, Appendix~\ref{app:validation}):
\begin{itemize}
  \setlength{\itemsep}{0pt}
  \item Wrong fails are infrequent and seldom recur, so a fresh grade corrects them: a single grade fails a working plan 0.2\% of the time, and a majority of five removes such isolated errors.\footnote{Measured before the audit fixes below, which changed no verdict on a working plan in a 169-plan check.}
  \item Wrong passes, in contrast, tend to recur, so we measured them with a second grader: GPT-6.1 Sol re-graded 477 plans that Opus had passed, and we adjudicated each disagreement by hand.
    Only 4 (0.8\%) were false, whereas Sol was wrong in 16 of the 20 disagreements.
  \item We planted 34 bugs, one at a time, in passing builds, 24 of them subtle, such as a total off by a cent; the grader caught every one (Table~\ref{tab:planted}).
  \item To find and fix the causes of wrong verdicts before release, we audited all 178 failed plans of Sol, Astra and Opus in 12 reference rounds (a round is one build of every product), and three authors manually verified the 50 builds that a retrospective review agent flagged as most ambiguous.
    The audit found 13 wrong verdicts and two points the specifications left unsaid; we fixed every one before release, without changing what any plan requires, and the other 159 failures are genuine bugs.\footnote{10 grader errors, 2 test defects and 1 harness fault; once the specifications were clarified, Sol, Astra and Opus pass both affected plans in all 24 independent builds (Appendix~\ref{app:validation}).}
  \item Review passes by agents and authors checked every stage against the real software and every plan for fairness, editing 117 of 571 draft plans without loosening any check; 27 were dropped by rule, leaving 544.
\end{itemize}
Each task received about 2.5 hours of review.

\subsection{Composition}
\label{sec:composition}

The eight public products contain 104 feature stages (9 to 14 each) and 252 test plans, and the nine held-out products 137 stages (12 to 18 each) and 292 plans.
Of the 544 plans, 463 (85\%) exercise features used together, 64 a single feature and 17 sign-in flows.
We release the eight public products with their PRDs, assets and test plans; the remaining nine are held out to guard against training on the tasks, and independent scores use all 17.

\section{Experimental Setup}
\label{sec:experiments}

This section describes the harness, the models we evaluate and the metrics we report.

\subsection{Harness}
Because many agent harnesses are tuned to their own company's models, to compare models rather than harnesses every builder runs under identical conditions, with the same system prompt, context compaction and model settings, in a configuration that mirrors how agentic coding platforms run agents in practice.\footnote{Medium reasoning effort, temperature 1.0, up to 300 steps and 2 hours per stage, and summarization past 200k tokens; only open-weight models' top-$p$ follows the developer (Appendix~\ref{app:setup}).}
This harness is deliberately minimal: one agent with a terminal, a file editor\footnote{The one provider-specific tool: GPT models edit with apply-patch, all others with string-replace.} and a task list, all stock tools of the OpenHands SDK, and far simpler than \platform{}'s production agent~\citep{zhong2026agent4}.
\platform{} also runs an internal benchmark variant on its production agent, whereas public scores use the open harness, so results are reproducible and no model gets a harness tuned to it.

\subsection{Models}
We evaluate six closed models, GPT-6.1 Sol, GPT-6 Astra, Opus 5.5, Sonnet 5.5, Grok 4.7 and Gemini 3.8 Flash, and three open-weight models, DeepSeek V4.1 Flash, Kimi K3 and GLM 5.3.
Each model builds every product four times, for 612 builds in total.
The independent runs cover all nine models and give every score in Table~\ref{tab:main}.

\subsection{Metrics}
\label{sec:metrics}
For each build, pass@1 equals 1 if every plan passes and 0 otherwise; it measures whether the agent delivered a product that works as specified.
The plan pass rate is the fraction of plans that pass.
A product that fails to start fails all of its plans.
Both scores are averaged first over a product's four builds and then over products, with 95\% intervals computed from the spread of each product's builds and an exact binomial range near 0\% (Appendix~\ref{app:grading}).
Cost is the builder's spend per build at list prices, and build time is the wall-clock time from the MVP's start to the last stage's end.

\section{Results}
\label{sec:results}

\begin{table}[t]
  \centering
  \small
  \setlength{\tabcolsep}{2.5pt}
  \begin{tabular}{@{}lrrrr@{}}
    \hline
    Model & pass@1 & Plan pass & Min. & Cost \\
    \hline
    GPT-6.1 Sol         & 64.7 $\pm$ 9.3  & 96.9 $\pm$ 1.2 & 178 & \$12.3 \\
    GPT-6 Astra         & 60.3 $\pm$ 10.4 & 98.0 $\pm$ 0.6 & 198 & \$89.9 \\
    Opus 5.5            & 48.5 $\pm$ 10.9 & 93.7 $\pm$ 4.2 & 73  & \$23.1 \\
    Sonnet 5.5          & 19.1 $\pm$ 9.8  & 79.2 $\pm$ 7.3 & 26  & \$4.4 \\
    Grok 4.7            & 1.5 (0--7.9)    & 52.9 $\pm$ 8.9 & 118 & \$63.9 \\
    DeepSeek V4.1 Flash & 0.0 (0--5.3)    & 77.4 $\pm$ 3.2 & 221 & \$2.5 \\
    GLM 5.3             & 0.0 (0--5.3)    & 66.0 $\pm$ 4.9 & 231 & \$27.8 \\
    Kimi K3             & 0.0 (0--5.3)    & 56.7 $\pm$ 8.1 & 221 & \$33.1 \\
    Gemini 3.8 Flash    & 0.0 (0--5.3)    & 38.1 $\pm$ 5.6 & 249 & \$27.0 \\
    \hline
  \end{tabular}
  \caption{The three strongest models pass 94 to 98\% of test plans but deliver a fully working product in only 49 to 65\% of builds.
    Each model built each of the 17 products four times in the independent runs, and scores average over these 68 builds.
    Values after $\pm$ and in parentheses are 95\% intervals.
    Time is in minutes and cost is in USD, both per build.}
  \label{tab:main}
\end{table}

\paragraph{Key results.}
As with $\tau$-bench's pass$^k$~\citep{yao2024taubench}, the headroom lies in reliability: the strongest models complete almost every product, but not consistently.
In Table~\ref{tab:main}, GPT-6.1 Sol achieves the highest pass@1, at 64.7\%, followed by GPT-6 Astra at 60.3\% and Opus 5.5 at 48.5\%; the intervals of Sol and Astra overlap.
These three models pass 93.7 to 98.0\% of their plans, and their failed builds fall only narrowly short: of their 84 failed builds that were seeded, 42 fail exactly one plan and 62 fail at most two.
Indeed, Astra has at least one passing build on 16 of the 17 products, Sol on 15 and Opus on 13; only on the product modeled on Figma do all 36 independent builds fail, mostly on its concurrent-editing and undo plans.
In contrast, none of the three open-weight models passes any of their 204 builds, although their plan pass rates still separate them.
DeepSeek V4.1 Flash, at 77.4\%, comes within two points of Sonnet 5.5 at about 60\% of its cost.

\subsection{Cost and Model Performance}
Build cost ranges from \$2.5 to \$89.9 and build time from 26 minutes to over four hours.
Astra costs about seven times as much per build as Sol for a pass@1 within five points.
Opus builds 2.4 times faster than Sol, in 73 minutes, with only 6\% fewer output tokens, pointing to serving speed.

\subsection{Model Performance over Versions}
\begin{table}[t]
  \centering
  \small
  \setlength{\tabcolsep}{2.5pt}
  \begin{tabular}{@{}lrrrr@{}}
    \hline
    Model & pass@1 & Plan pass & Min. & Cost \\
    \hline
    GPT-6.1 Sol$^\ast$ & 53.1 & 94.9 & 170 & \$11.8 \\
    GPT-6.1 Terra      & 46.9 & 94.5 & 122 & \$6.2 \\
    GPT-6 Sol          & 31.2 & 84.0 & 88  & \$19.4 \\
    GPT-5.6 Sol        & 3.1  & 67.5 & 73  & \$26.5 \\
    GPT-6 Luna         & 0.0  & 59.8 & 188 & \$1.7 \\
    GPT-5.5            & 0.0  & 51.5 & 46  & \$30.1 \\
    GPT-5.4            & 0.0  & 41.5 & 78  & \$22.9 \\
    GPT-5.6 Terra      & 0.0  & 26.1 & 39  & \$5.3 \\
    GPT-5.6 Luna       & 0.0  & 6.6  & 35  & \$0.5 \\
    \hline
  \end{tabular}
  \caption{Newer GPT releases pass more test plans, but pass@1 rises only once the plan pass rate exceeds about 70\%.
    Each model built the eight public products four times.
    Columns follow Table~\ref{tab:main}.
    The starred row is the external team's run; the others are ours.}
  \label{tab:versions}
\end{table}

We also built the eight public products four times each with GPT releases from GPT-5.4 to GPT-6.1, taking GPT-6.1 Sol from the independent runs (Table~\ref{tab:versions}).
With each release, the strongest model's plan pass rate rises steadily, from 41.5\% to 51.5, 67.5, 84.0 and 94.9\%, while its pass@1 stays near zero through GPT-5.6, then rises to 31.2 and 53.1\%.
The two metrics therefore complement each other: the plan pass rate separates releases whose pass@1 is still near zero, such as the two Luna releases, and pass@1 separates the strongest.
GPT-6.1 Sol, best on both, costs \$11.8 per build, down from a peak of \$30.1 for GPT-5.5, and GPT-6.1 Terra, at \$6.2, exceeds all earlier releases.

\subsection{Failure Analysis}
\paragraph{Where builds fail.}
In the independent runs, failures are not concentrated where features interact: for Opus, plans that exercise features together fail about as often as single-feature plans (6.4\% against 6.2\%), so the 74\% of failed Sol, Astra and Opus builds that fail only multi-feature plans reflects the 85\% share of such plans.
The difficulty lies in breadth instead: Opus passes 93.7\% of plans, but since every plan must pass after the final stage, fewer than half of its builds pass.

\paragraph{Complete failures.}
Builds that fail every plan are rare: 2 of the 204 independent builds of Sol, Astra and Opus fail every plan, both from Opus, because seeding failed on every plan.
In our reference runs, every build that failed in this way had schema code that used a table before the statement that creates it, and moving one to four SQL statements made each application start.
Every builder is told its product first runs on an empty database, yet a clean install reveals each bug.

\paragraph{Visual quality.}
Although plans assess functionality rather than visual design, the builds differ markedly in appearance.
Among the four models in Figure~\ref{fig:shots} (Appendix~\ref{app:shots}), Sonnet is the outlier: on the same seeded file browser, it renders a plain list with a truncated button, whereas Sol, Astra and Opus produce polished, product-like pages.
Sonnet also writes roughly half as many output tokens per build (0.20 million versus 0.36 to 0.39 million).
Under a minimal prompt, Sonnet invests little in aesthetics; on \platform{}, however, more detailed instructions elicit polished designs from it, so a platform serving this model can recover much of the gap through its own system instructions.

\section{Deployment and Lessons}
\label{sec:practice}

\platform{} uses ViBench 2.0 for the deployment decisions below, and Fireworks includes it in its Specialized Intelligence Index~\citep{fireworks2026sii}.
Beyond the score, \platform{} examines ViBench 2.0 builds with the trace-analysis tooling it applies in production, which samples, clusters and monitors agent trajectories~\citep{furman2026closing} to detect harness faults and to study visual quality and adherence to visual instructions.
We distill six lessons.

\paragraph{Two metrics answer different questions.}
\platform{}'s core loop, the main agent, generally runs on the models with the highest pass@1, since users expect a fully working product.
For specialized workflows where speed and cost matter more, a cheaper model at a faster service tier, whose plan pass rate stays close to the leaders', is often adopted.
Candidates that fell short were rejected, including one for the core loop.

\paragraph{Context handoff matters over long horizons.}
Because each request builds on context from earlier ones, what survives compaction determines what later requests can build on, so compaction is a first-order concern for any agent that works over long horizons: hill-climbing on an internal variant of ViBench 2.0 guided how \platform{} optimized it.

\paragraph{The plan pass rate moves first.}
Across versions of a model family, the plan pass rate rises before pass@1 (Section~\ref{sec:results}), so it is the earlier signal.

\paragraph{Platforms should check deployments.}
Agents do not reliably test that their product deploys successfully (Section~\ref{sec:results}); platforms should validate it.

\paragraph{Fair, difficult tests need repeated review.}
Review passes edited about a fifth of the draft plans without loosening any check (Section~\ref{sec:validation}).

\paragraph{Both kinds of grading error must be measured.}
Re-grading removes wrong fails but not wrong passes, so each needs its own measurement.

\section{Conclusion}
\label{sec:conclusion}
We introduced ViBench 2.0 to measure whether an agent can build a fully working product over a long horizon.
The strongest models implement nearly every feature, yet at least a third of their builds still fail one or more plans.
Yet each release advances the cost--performance frontier: across three Sol releases, pass@1 on the public products rises from 3\% to 53\% while the cost per build falls from \$27 to \$12.
We expect open-weight models to follow, and will keep adding to and extending the benchmark so that it remains unsaturated; the remaining gap is also an engineering target for production agents such as \platform{}'s.

\section*{Limitations}

\textbf{Contamination.}

Nine products are held out from public release and shared only with the external team, under a non-disclosure agreement that prohibits training on them. The eight public products could be used for training, so every public file carries a canary string for filtering. Because independent scores include the held-out products, members of the public can reproduce only the public subset, which we report separately (Appendix~\ref{app:products}).
The public subset is nonetheless a faithful proxy, and a harder one.
The three strongest models score 7 to 12 points lower on it in pass@1, leaving even more headroom, while it nearly preserves the ranking of all nine models (Spearman $\rho = 0.99$) and its plan pass rates closely track those on the held-out products ($\rho = 0.97$).
We judge holding out products worth this cost, since tasks that no model can train on keep the comparison fair and discourage optimizing for the benchmark over building products.

\textbf{Grading.} 
Passed plans are not graded again, so a wrong pass would stand; a second grader puts such errors at only 0.8\% of passed plans.
Our grader studies used our reference runs and the grader as it stood before the audit fixes, which changed no verdict on a working plan.
The grader is a closed model that is also evaluated as a builder and may be deprecated, though we found no sign that it favors its own family (Appendix~\ref{app:validation}).

\textbf{Harness.} 
All builders run in a single agent harness built on the OpenHands SDK; a model tuned for a different agent setup, including a production agent such as \platform{}'s, may therefore score differently.

\textbf{Scope.} 
ViBench 2.0 covers full-stack web products backed by a PostgreSQL database, specified in English and modeled on 17 widely used products.
It evaluates functional correctness only: visual design is subjective and difficult to grade automatically, and neither the speed nor the cost of the finished product is measured.
Every model also builds the features in one fixed order and, unlike in real use, receives no bug reports between requests, so all models receive the same requests.

\textbf{Cost.} 
Grading a full round, one build of each of the 17 products, with Opus 5.5 as the grader costs about \$1,440, or \$85 per build, comparable to running the most expensive builder in Table~\ref{tab:main}.
The public subset halves this, at about \$670 per round, while remaining harder and nearly preserving the ranking of models (see Contamination).
We are actively developing a lighter benchmark variant to cut the cost further, so that more teams can run it.

\textbf{Variance.} 
With four builds per product, pass@1 has 95\% intervals of about $\pm$10 points, too wide to separate models as close as Sol and Astra; the plan pass rate, at about $\pm$1 point for Sol and Astra, gives a finer comparison.

\textbf{Reproducibility.} 
We invite the community to report any plan that falls short of the criteria in Section~\ref{sec:validation}, as our audits may not have caught every case.

\section*{Ethical Considerations}
Each product is modeled on publicly available software for evaluation only; the product names are trademarks of their owners, who are not affiliated with ViBench 2.0.
The benchmark contains no user data: we wrote every PRD and test plan ourselves, and all human review in our audits was done by the authors.
To keep the benchmark's scores meaningful as models improve, we also guard against training on the tasks.
We do not train on the nine held-out products; they are shared only with the external team, under a non-disclosure agreement, and our model-provider partners use the benchmark only for evaluation and not as training data.
All runs used zero-data-retention terms, so model providers do not keep the requests for training, and we did not share build or grading traces with model developers.
The public products carry a canary string so that they can be filtered from training data.
We work with some of the providers whose models we evaluate; to compare models fairly, every model runs in the same harness with the same prompt and limits.
Three further caveats remain: the grader, Opus 5.5, also builds, and Appendix~\ref{app:validation} finds no sign that it favors its own family; open-weight costs use Fireworks list prices; and the version study covers one provider only.

\section*{Acknowledgments}

We thank the Fireworks AI team\footnote{\url{https://fireworks.ai}}, in particular
Vivek Chauhan and Charlie Gu for leading the partnership that brought ViBench to
the Specialized Intelligence Index (SII), and Sandeep Singh and Arleen Liu for conducting the independent runs. We also thank
Georgian\footnote{\url{https://georgian.io}}, our collaborator on ViBench since its
inception, and in particular Kartik Gupta for building the Harbor adapter, Aditya
Shukla and Scott Cui for their support, and Pashootan Vaezipoor for shaping the
project's direction. Finally, we thank the rest of the AI team at Replit for their
contributions to ViBench, and the anonymous reviewers for their helpful feedback.

\bibliography{custom}

\appendix

\section{Related Work}
\label{app:related}

Most coding-agent benchmarks work inside a codebase that already exists.
SWE-bench~\citep{jimenez2024swebench} asks agents to resolve GitHub issues, SWE-Bench Pro~\citep{deng2025swebenchpro}
and SWE-EVO~\citep{le2025sweevo} stretch this to tasks that take engineers hours to days and to whole release
transitions, and FeatureBench~\citep{zhou2026featurebench} and FeatBench~\citep{chen2025featbench} ask for new
features, so the stack and architecture are chosen before the agent arrives.
Benchmarks that build from a specification leave more open: Commit0~\citep{zhao2025commit0} has agents fill in a
stubbed library, NL2Repo-Bench~\citep{ding2026nl2repobench} has them write one from a requirements document,
BaxBench~\citep{vero2025baxbench} has models write a backend in a given framework, and
ProgramBench~\citep{yang2026programbench} has agents rebuild a command-line program from its executable and
documentation alone.
Terminal-Bench~\citep{merrill2026terminalbench} covers hard tasks of many kinds in a terminal, and its Harbor format
is the one ViBench 2.0 uses.
These benchmarks are built to stay hard, since benchmarks tend to saturate within a few years of
release~\citep{akhtar2026plateau}, but each one grades a single pass over a fixed target.
ProjectEval~\citep{liu2025projecteval}, E2EDev~\citep{liu2026e2edev}, WebGen-Bench~\citep{lu2025webgenbench},
VISTA~\citep{guo2026vista}, Vibe Code Bench~\citep{tran2026vibecodebench} and ViBench~\citep{zhong2026vibench} go
further, building whole applications and checking them through the interface.

Many of these benchmarks grade with an agent that operates the application through a browser or GUI, among them
WebGen-Bench~\citep{lu2025webgenbench}, Vibe Code Bench~\citep{tran2026vibecodebench},
RealDevWorld~\citep{bian2025realdevworld}, WebCraftBench~\citep{liu2026webcraftbench} and
ViBench~\citep{zhong2026vibench}.
Agent-as-a-Judge~\citep{zhuge2025agentasajudge} instead reads the workspace and
ArtifactsBench~\citep{zhang2025artifactsbench} judges screenshots, and most of these graders are validated against
human labels.
These graders err in both directions.
LLM judges can favor their own outputs or model family~\citep{zheng2023judging,panickssery2024llm,bodhwani2026gauge},
change verdicts when grading the same input again~\citep{haldar2025ratingroulette} and trail experts on web
applications~\citep{li2026webdevjudge}.
Browser agents hunting for bugs without a checklist miss more than half of them~\citep{hong2026cat}, and scripted
checks both fail successful runs~\citep{dong2026misscore} and pass incorrect ones~\citep{yu2025utboost,zhu2025abc}.
ViBench 2.0 uses the browser-agent grader of ViBench~\citep{zhong2026vibench}, but re-grades each failed plan until
three of up to five grades agree and discards grades without on-screen evidence.
It checks the grader with repeated grades and planted bugs and, like EvoWeb~\citep{nukrai2026evoweb}, with a
cross-provider audit and a same-family bias test.

Most of these benchmarks also ask for at most one change after the first build, while real products grow feature by
feature on code the agent wrote earlier.
In ViBench~\citep{zhong2026vibench}, most models already do worse when they extend their own build than a reference
one.
Among multi-turn benchmarks, Asuka-Bench~\citep{wang2026asukabench}, SWE-Together~\citep{wu2026swetogether} and
ICAE-Bench~\citep{peng2026icaebench} refine a fixed target with a simulated user.
Web-Bench~\citep{xu2025webbench}, SlopCodeBench~\citep{orlanski2026slopcodebench},
StaminaBench~\citep{sobal2026staminabench} and EvoCode-Bench~\citep{shen2026evocodebench} grow software and score
each step as it is completed, and CodeClash~\citep{yang2026codeclash} has agents revise their code over many rounds toward a
fixed goal.
None of them waits until the last request to judge the whole product the way its user would.

The benchmarks closest to ViBench 2.0 grow web applications over many requests or build whole products, but each
stops short.
MACBench~\citep{gui2026macbench} follows frontends through changing requirements and finds agents failing on
application logic, and ProductWebBench~\citep{wu2026productwebbench} finds them breaking states they had already
passed.
EvoWeb~\citep{nukrai2026evoweb} builds and extends an app over ten user-level requests in one conversation and tests
every version with a browser agent, but its apps are client-side only.
Vibe Code Bench 1-100~\citep{tran2026vcb1100} grows full-stack applications but starts from one that already passes
its tests and stops at the first failure.
SWE-Marathon~\citep{desai2026swemarathon} clones products such as Slack and Mastodon, each from
a single specification.

ViBench 2.0 addresses the remaining gap: long-horizon, greenfield agentic engineering of a whole product, judged as a whole after the last request.

\section{Benchmark Details}
\label{app:benchmark}

\paragraph{PRDs and assets.}
Where behavior depends on constants, such as the ride-hailing product's fares and fees, the PRD fixes them in an environment variable so that tests can check exact amounts.
Some MVP PRDs also come with assets, such as a product catalog, a brand logo, a list of postcodes, or a template for the configuration the app reads at startup.
Assets let us simplify features that would otherwise rely on outside services, such as a map, so that flaky external APIs cannot affect grading.

\paragraph{Feature dependencies.}
For each product, we first wrote one PRD covering its full feature set, then split it into an MVP and one feature per stage.
To order the stages, we drew a dependency graph for each product, in which an edge from feature A to feature B means that B builds on behavior that A introduces (Figure~\ref{fig:deps}).
We then ordered the stages topologically, breaking ties at random, so that every feature arrives after the features it builds on.

In the product modeled on Uber (Figure~\ref{fig:deps}), for instance, earlier stages let drivers own several cars, each in its own tier, and charge riders a fee for cancelling once a driver has the ride.
A later stage shows each driver's earnings per trip, which must name the car driven and include those fees.
The last stage lets riders split a fare without changing what the driver earns.
Plans that test features together are not limited to the graph's edges.
The plan that checks those earnings, for instance, also meets family profiles, scheduled rides, stops and declined offers, each through a different step.

\paragraph{Test plans.}
Many plans name the mistakes they are written to detect, such as a build that splits the fare but not the fee.
To keep one bug from costing more than one step, each behavior is graded by a single plan, and other plans that run into it only note what happened; within a plan, each step is graded against what actually happened before it, so one failure does not spread.
Plans also spell out the variants they accept, such as status labels worded differently from the PRD, so that each failed plan points to something the builder was told to do and did not do.

\paragraph{Environment.}
Every stage's system prompt asks for the app's own username-and-password sign-in, rather than an outside sign-in service, and UTC as the default time zone.
A stage that runs out of steps ends there and the next one begins; one that fails in any other way, such as running out of time, stops the build, and the product is graded as it stands.
Infrastructure failures, such as a provider outage, are retried and never counted against the model.
Builds are long: in the independent runs, a build of Sol, Astra or Opus reads 55 to 63 million input tokens, almost all of them cached, and writes 0.36 to 0.39 million output tokens, and summarization fired in every reference build of Sol and Astra and in 67 of 68 of Opus.
Every build, seed and grade container has 4 CPUs and 8 GB of memory and runs Python 3.12 and Node.js 22 on Debian, beside a PostgreSQL 17 database.

\section{Grading Details}
\label{app:grading}

\begin{figure*}[h]
  \centering
  \colorlet{vbPass}{vbgreen}
\colorlet{vbFail}{vbaccent}
\colorlet{vbOut}{vbmuted}
\colorlet{vbRetry}{vbmuted}
\begin{tikzpicture}[
    x=1cm, y=1cm,
    sbox/.style={fill=vbfill, rounded corners=4pt, font=\scriptsize, align=center, inner sep=2pt,
      text width=1.7cm, minimum height=1.05cm},
    send/.style={rounded corners=2.6mm, font=\footnotesize, inner xsep=4pt, inner ysep=3pt,
      minimum height=0.52cm, text depth=0pt, line width=0.7pt},
    spass/.style={send, fill=vbgreen!12, draw=vbgreen, text=vbink},
    sfail/.style={send, fill=vbpale!55, draw=vbaccent, text=vbink},
    sout/.style={send, fill=vbfill, draw=vbmuted, text=vbmuted},
    sbadge/.style={circle, inner sep=0pt, minimum size=3.4mm, draw=none},
    smark/.style={line width=0.9pt, draw=white, line cap=round, line join=round},
    sflow/.style={-{Stealth[length=4pt, width=3.5pt]}, line width=0.5pt, draw=vbdark},
    sretry/.style={sflow, draw=vbRetry},
    slab/.style={font=\scriptsize, align=center, inner sep=1pt},
  ]
  \newcommand{\sbadge}[3]{%
    \node[sbadge, fill=#2] (#1b) at ([xshift=2.6mm]#1.west) {};
    \ifx#3c\draw[smark] ([xshift=-0.85mm]#1b.center) -- ([xshift=-0.2mm, yshift=-0.7mm]#1b.center) -- ([xshift=0.95mm, yshift=0.75mm]#1b.center);\fi
    \ifx#3x\draw[smark] ([xshift=-0.7mm, yshift=-0.7mm]#1b.center) -- ([xshift=0.7mm, yshift=0.7mm]#1b.center)
      ([xshift=-0.7mm, yshift=0.7mm]#1b.center) -- ([xshift=0.7mm, yshift=-0.7mm]#1b.center);\fi
    \ifx#3d\draw[smark] ([xshift=-0.75mm]#1b.center) -- ([xshift=0.75mm]#1b.center);\fi}
  \coordinate (origin) at (0, 0);
  \coordinate (r2) at (0, -1.9);
  \coordinate (sd) at (0, -0.95);

  \node[slab, anchor=west, text width=1.45cm] (in) at (origin) {finished app, one test plan};
  \node[sbox, anchor=west] (write) at ([xshift=4mm]in.east) {\textbf{Write seed}\\ seeding agent};
  \node[sbox, anchor=west] (check) at ([xshift=5.5mm]write.east) {\textbf{Check seed}\\ replay on an empty database};
  \node[sbox, anchor=west] (start) at ([xshift=14mm]check.east) {\textbf{Start app}\\ new containers};
  \node[sbox, anchor=west] (grade) at ([xshift=11mm]start.east) {\textbf{Grade}\\ browser agent};
  \node[spass, anchor=west] (pass) at ([xshift=15mm]grade.east) {\hspace{3.6mm}Plan passes};
  \draw[sflow] (in) -- (write);
  \draw[sflow] (write) -- (check);
  \draw[sflow] (check) -- node[slab, above] {seed passes} (start);
  \draw[sflow] (start) -- node[slab, above] {app up} (grade);
  \draw[sflow, draw=vbPass] (grade) -- node[slab, above] {passes} (pass);

  \draw[sretry, rounded corners=3pt] (check.north) -- ++(0, 0.4) -|
    node[slab, above, pos=0.25] {app silent after replay\\\textcolor{vbRetry}{seed retried 1$\times$}} (write.north);
  \draw[sretry] ([xshift=3mm]write.south) to[out=-70, in=-110, looseness=4]
    node[slab, below] {seeding run crashed\\\textcolor{vbRetry}{retried 2$\times$}} ([xshift=-3mm]write.south);
  \draw[sretry] ([xshift=3mm]start.north) to[out=70, in=110, looseness=4]
    node[slab, above] {app did not start\\\textcolor{vbRetry}{retried 2$\times$}} ([xshift=-3mm]start.north);

  \node[sfail] (seedfail) at (check |- r2) {\hspace{3.6mm}Plan fails};
  \node[sout] (leftout) at (start |- r2) {\hspace{3.6mm}Plan left out};
  \node[sbox, minimum height=0.8cm] (regrade) at (grade |- r2) {\textbf{Re-grade}\\ majority decides};
  \node[sfail] (fail) at (pass |- r2) {\hspace{3.6mm}Plan fails};

  \sbadge{pass}{vbgreen}{c}
  \sbadge{seedfail}{vbaccent}{x}
  \sbadge{fail}{vbaccent}{x}
  \sbadge{leftout}{vbsilver}{d}
  \draw[sflow, draw=vbFail] (check) -- node[slab, right] {seed still fails} (seedfail);
  \draw[rounded corners=3pt, line width=0.5pt, draw=vbdark] ([xshift=4mm]start.south) |-
    node[slab, above, pos=0.75] {still down} (grade |- sd);
  \fill[vbdark] (grade |- sd) circle (1.2pt);
  \draw[sflow] (grade) -- node[slab, left, pos=0.62] {fails, or thrown out} (regrade);
  \draw[sflow, draw=vbPass] (regrade.north east) -- node[slab, below right, pos=0.55] {majority pass} (pass.south west);
  \draw[sflow, draw=vbFail] (regrade) -- node[slab, above] {majority fail} node[slab, below] {\textcolor{vbRetry}{or tie}} (fail);
  \draw[sflow, draw=vbOut] (regrade) -- node[slab, above] {no grade} node[slab, below] {\textcolor{vbRetry}{finishes}} (leftout);
  \draw[sretry] ([xshift=3mm]regrade.south) to[out=-70, in=-110, looseness=4] ([xshift=-3mm]regrade.south);
  \node[slab, anchor=north west, align=left] at ([xshift=5mm, yshift=-1mm]regrade.south)
    {one more grade until\\\textcolor{vbRetry}{3 of up to 5 agree}};
\end{tikzpicture}
  \caption{Seeding and grading one test plan.
    Grey loops repeat a step, and each retry count is the most that step is retried.
    A plan that passes its first grade is not graded again, and a thrown-out grade (Section~\ref{sec:protocol}) does not count toward the majority.}
  \label{fig:seed-grade}
\end{figure*}
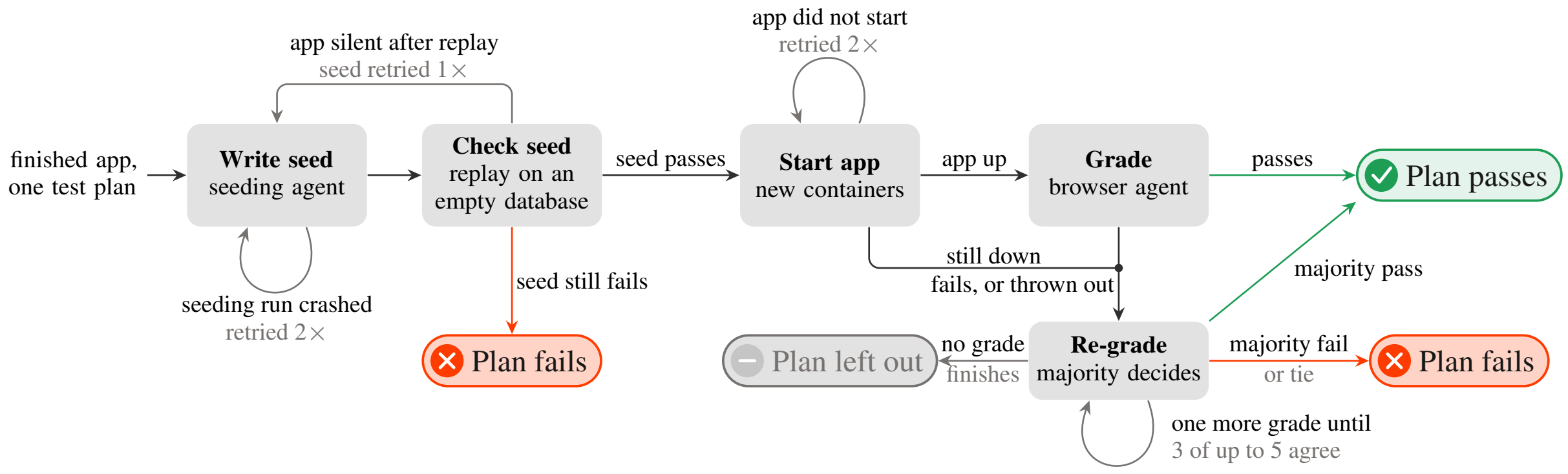

\paragraph{Seeding.}
Some plans need data in place before they start, such as accounts and records that already exist.
Of the 252 public plans, 93 (37\%) need this setup: 32 create records, and 61 only set configuration, such as a product's fares.
For each such plan, the seeding agent reads the finished app and writes a script that creates this data, using any seed files that come with the plan.
Every grade starts from a fresh container and an empty database and runs this script first, so all grades of a plan start from the same state.
A plan whose seeding fails counts as failed.

\paragraph{Grader rules.}
The grader receives one plan and the rules that all plans share.
It acts as a human tester would: it follows the plan's steps in order through the browser only, with no API calls, database writes, changes to the page's code, or reloads that the plan does not allow.
For each step it must quote, from its own browser output, what it saw on screen; API responses, the database and the source code never decide a check.
A step earns its points only if every check in it passes, and a plan passes only if every step does.
The grader sees the screenshot and accessibility snapshot of its two newest browser steps in full; older page images are replaced, eight at a time, by a note listing their saved files, while each step's script and result are kept.
It clicks, types key by key and drags; when a plan has several users, it acts in a separate browser session for each; after each action it observes a screenshot and an accessibility snapshot of the page; and it may consult the application's code to navigate, though the code never decides a check.
Each grade may take up to 1 hour.

\paragraph{Discarded grades.}
A discarded grade does not count toward the five.
If none of a plan's grades finishes after every retry, its result is \emph{no grade}.
This is rare: in the independent runs, only about 30 of 19,584 plans (0.2\%) ended this way, 24 of them from Gemini 3.8 Flash.
By design, the scorer treats such a result as an infrastructure fault rather than a defect in the build: it leaves the plan out of both scores and lists it for audit, so a build that passes all of its other plans still counts as passing.
In the independent runs, every affected build also fails another plan, so these results changed no pass@1 score.

\paragraph{Cost.}
In our reference runs, seeding and grading one 17-product round with Opus 5.5 at medium effort, including re-grades, cost \$1,509 at 571 plans, or about \$1,440 at the final 544 (\$2.64 per plan), about \$85 per build.

\paragraph{Intervals.}
For each product $a$ with builds $r$, the score is the mean over products of each product's mean over builds.
The 95\% interval is $t_{0.975,\,n-k}\sqrt{\sum_a s_a^2/n_a}\,/\,k$, where $k$ is the number of products, $n$ the number of builds, $n_a$ and $s_a^2$ the number and sample variance of product $a$'s builds.
For 17 products with 4 builds each, $t_{0.975,51} = 2.01$.
Near 0\% or 100\%, where this interval would cross the limit, we report the exact binomial (Clopper--Pearson) range instead.

\section{Grader Validation Details}
\label{app:validation}

\begin{table}[t]
  \centering
  \small
  \setlength{\tabcolsep}{3pt}
  \begin{tabular}{@{}p{0.25\columnwidth}p{0.72\columnwidth}@{}}
    \hline
    Test & Result \\
    \hline
    Consistency & 99.0\% of 97 plans receive the same verdict in 10 of 10 grades. \\
    Wrong fails & 3 of 1,521 single grades fail a working plan (0.20\%; 0.07--0.58\%). \\
    Wrong passes & 4 of 477 passes are false (0.8\%; 0.3--2.1\%), settled by hand after re-grading with Sol. \\
    Planted bugs & 0 of 96 grades pass a plan while its bug shows (0--3.9\%). \\
    Same-family bias & 4 of 237 wrong passes on Claude-built apps, 0 of 240 on GPT-built apps; within chance. \\
    Failure audit & Every issue found in 178 failed plans of Sol, Astra and Opus was fixed before release; the other 159 are real bugs. \\
    \hline
  \end{tabular}
  \caption{Grader validation on our reference runs.
    Ranges are 95\% intervals.
    Wrong fails are single grades, before the majority of five.
    All studies used the grader as frozen before the audit fixes.}
  \label{tab:grader}
\end{table}

\paragraph{Choosing the grader.}
We graded every plan of the same two builds of the product modeled on Uber 10 times with each candidate grader, 640 grades each.
Opus 5.5 at medium effort made the fewest wrong fails (12, 1.9\%), ahead of GPT-6.1 Sol at low effort (24, 3.8\%), Opus 5.5 at low effort (26, 4.1\%) and Sol at medium effort (29, 4.5\%).
On 2,176 plans from an earlier set of builds, Sol at low effort returned 171 wrong verdicts (7.9\%) and Opus 17 (0.8\%).

\paragraph{Consistency and wrong fails.}
We graded three builds 10 times each, and 99.0\% of their 97 plans received the same verdict every time.
Across these grades and 291 more plans drawn at random from all 16 reference rounds, each graded twice more, a single grade failed a working plan 3 times in 1,521 (0.20\%, 95\% interval 0.07--0.58\%).
In the 947 grades of the 10-grade test in which the app worked, the grader failed none (95\% upper bound 0.4\%).
A majority of five removes such isolated errors; the wrong fails that survived it in our audit had a cause in the plan's text or in the harness that recurred on every grade, and we fixed each one (Section~\ref{sec:validation}).

\paragraph{Wrong passes.}
A second grader from another provider, GPT-6.1 Sol, graded again 477 plans that the ViBench 2.0 grader had passed, up to two per product in each of the 16 reference rounds.
We settled each disagreement by hand.
Our grader wrongly passed 4 of 477 (0.8\%, 95\% interval 0.3--2.1\%), and Sol was wrong in 16 of 20 disagreements.

\paragraph{Planted bugs.}
We planted 34 bugs, one at a time, in builds that passed the plan aimed at each bug, and graded that plan three times; we count the 96 grades in which the bug was visible, leaving out one discarded grade, two in which the grader never reached the bug and the three grades of one obvious bug that the interface does not show.
24 of the bugs are subtle: a total off by one cent, a change that lasts only until the page reloads, a value silently dropped, a refusal with no message, or an action that fires early (Table~\ref{tab:planted}).
The grader never passed a plan while its bug was visible: 0 of 96 grades (95\% interval 0--3.9\%).
On the same builds without the bugs, it wrongly failed 0 of 24 grades.

\begin{table}[t]
  \centering
  \small
  \setlength{\tabcolsep}{3pt}
  \begin{tabular}{@{}lr>{\raggedright\arraybackslash}p{0.6\columnwidth}@{}}
    \hline
    Bug kind & Passed & Example \\
    \hline
    Cent off & 0/15 & An \$11.11 ride pays the driver \$8.34, not \$8.33 (Uber). \\
    Reverts & 0/15 & A renamed dropdown option looks saved but reverts on reload (Asana). \\
    Dropped & 0/15 & A one-letter typo on the server discards every edited task description (Asana). \\
    No refusal & 0/14 & Cancelling a started ride fails without a message (Uber). \\
    Timing & 0/10 & A scheduled ride enters dispatch 45 seconds early (Uber). \\
    Obvious & 0/27 & Story points accept 4, 7 and 20, off the Fibonacci scale (Jira). \\
    \hline
  \end{tabular}
  \caption{The grader caught every planted bug.
    Passed counts the grades that passed the plan aimed at a visible bug.
    Each kind has four or five bugs; obvious bugs have ten.
    Bugs were planted in our reference builds; examples come from the public products.}
  \label{tab:planted}
\end{table}

\paragraph{Failure audit.}
The audit of Section~\ref{sec:validation} traced 178 failed plans of Sol, Astra and Opus in 12 reference rounds.
It found 13 wrong verdicts: 10 grader errors, such as a drag that picked up the wrong card, 2 test defects, such as a plan that expected an edit from a stale page to succeed, and 1 harness fault, a browser script sent twice after a dropped connection.
After each was fixed and re-graded, 159 of the 165 remaining failures are genuine bugs.
The other 6 turned on two points the specifications left unsaid: whether a cancelled order that never shipped reads as fulfilled (one plan, five builds), and whether a removed editor's past work stays credited to her (one build).
We clarified both specifications and the plans that check them before release, and the clarifications worked: in the independent runs, Sol, Astra and Opus pass both affected plans in all 24 of their builds.

\paragraph{Same-family bias.}
Opus 5.5 grades every builder, including itself and Sonnet 5.5, and we found no case of it passing a visible defect in a Claude-built app.
In the wrong-pass audit, it wrongly passed 4 of 237 plans on Claude-built apps and 0 of 240 on GPT-built apps.
In two of the four it never saw the bug, the other two were borderline calls, and with four cases the gap is within chance (95\% interval $-0.2$ to 4.3 points).

\paragraph{Builds that do not start.}
Every reference build that did not start uses a table or column before the statement that creates it; because each app runs its schema as one query, the first error rolls back every table.
We reproduced each failure on an empty PostgreSQL database, and moving one to four SQL statements makes each app start.

\section{Experiment Setup Details}
\label{app:setup}

GPT models edit files with an apply-patch tool and the others with a string-replace editor, following each provider's recommendation, as in ViBench~\citep{zhong2026vibench}.
Open-weight models use the top-$p$ their developers recommend for agentic use (1.0 for Kimi K3 and 0.95 for GLM 5.3 and DeepSeek V4.1 Flash), and their provider keeps the reasoning the builder sends back.
Each model's output and context limits follow its provider's.
Closed models are called through their providers' APIs and open-weight models through Fireworks, at its list prices.
The independent builds ran on release 2.0.1.beta and the runs of earlier GPT releases (Section~\ref{sec:results}) on 2.0.5; both use the same products, plans and model settings, as their matching dataset hashes confirm, and ViBench 2.0.6 keeps all three.
These settings match the release of the independent runs; later releases changed only how cost and intervals are reported.

\paragraph{Independent runs.}
The external team ran the public release without modifying it: after supplying API keys and machine settings, it ran one command per model, which builds, seeds, grades, confirms and scores every product.
Each run records its provenance, namely the harness commit, the dataset hash, the container image and the models, and the scorer does not pool runs whose settings differ; every independent run records release 2.0.1.beta.
A model provider or platform can therefore reproduce any score on the public products, or place a new model beside the published ones, without involving the authors.

\paragraph{Our runs.}
We also ran the full benchmark four times with Sonnet 5.5 and Opus 5.5 ourselves.
Our Sonnet builds reach 19.1 pass@1 and 77.1 plan pass rate, against the external team's 19.1 $\pm$ 9.8 and 79.2 $\pm$ 7.3, and our Opus builds reach 54.4 $\pm$ 9.5 and 90.5 $\pm$ 5.2, against 48.5 $\pm$ 10.9 and 93.7 $\pm$ 4.2, well within each other's intervals.

\section{Build Screenshots}
\label{app:shots}

\begin{figure*}[h]
  \centering
  \begin{tabular}{@{}cc@{}}
    \includegraphics[width=0.48\textwidth]{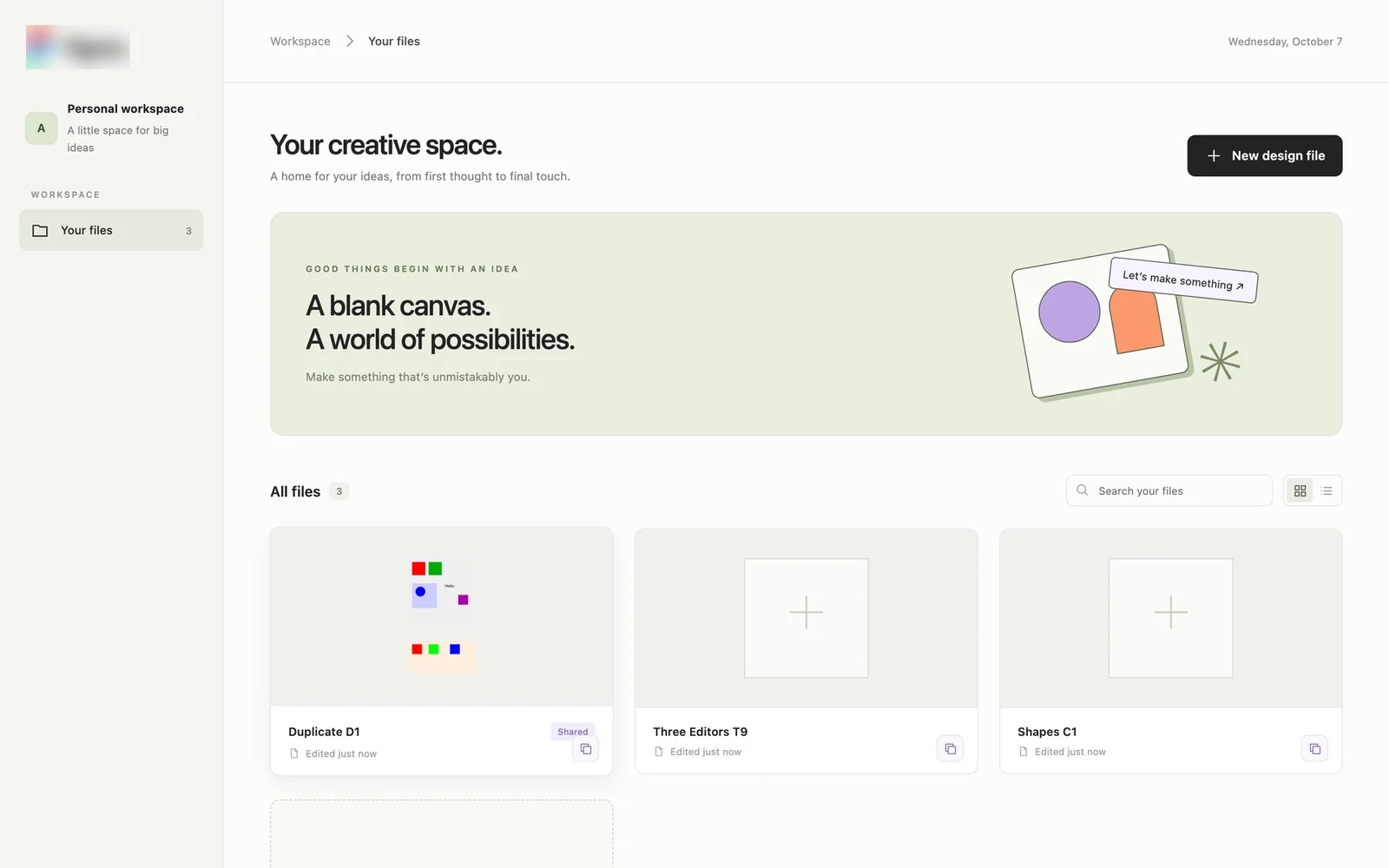} &
    \includegraphics[width=0.48\textwidth]{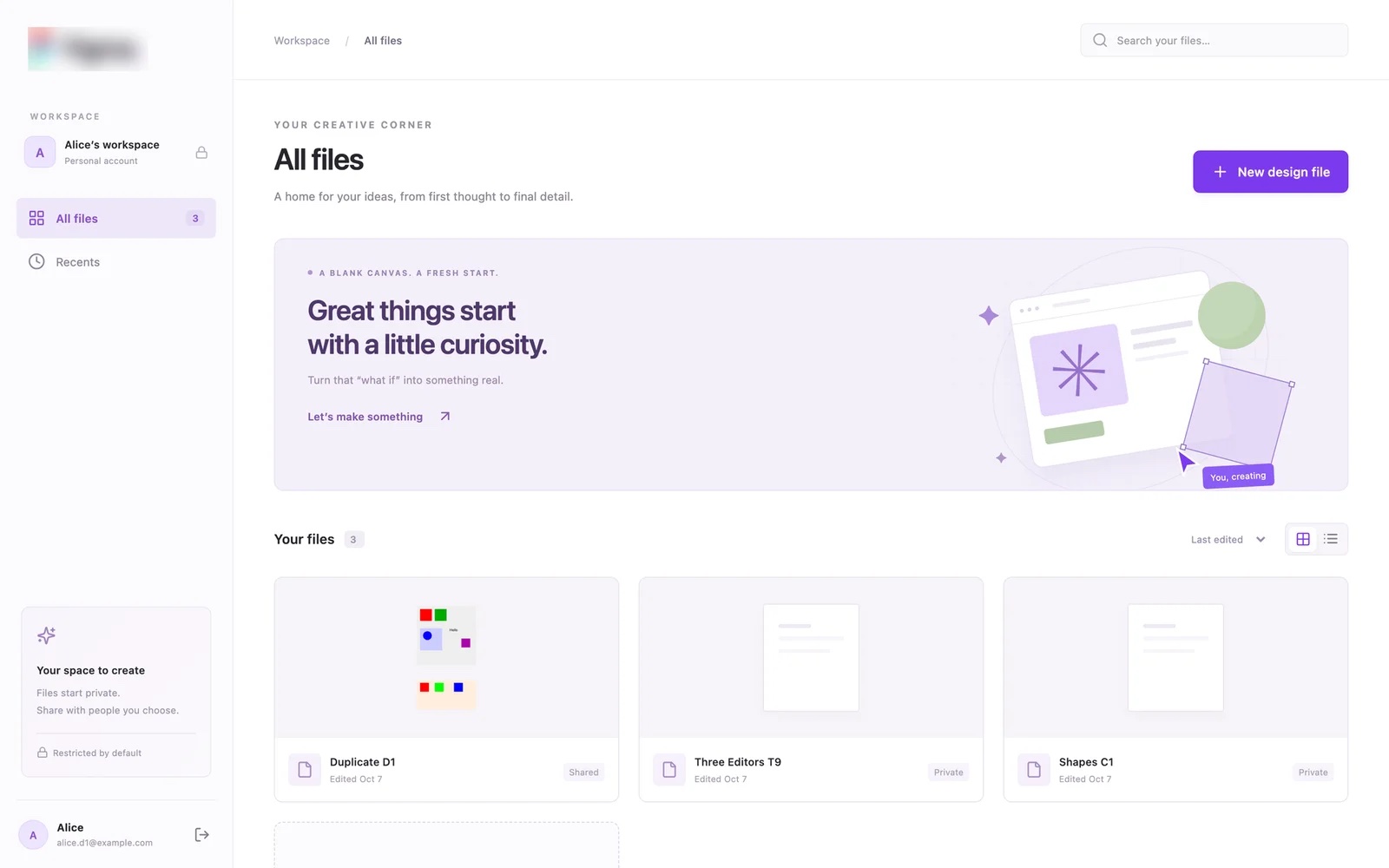} \\
    \small (a) GPT-6.1 Sol, build 4: passes every plan & \small (b) GPT-6 Astra, build 1: fails 1 plan \\[4pt]
    \includegraphics[width=0.48\textwidth]{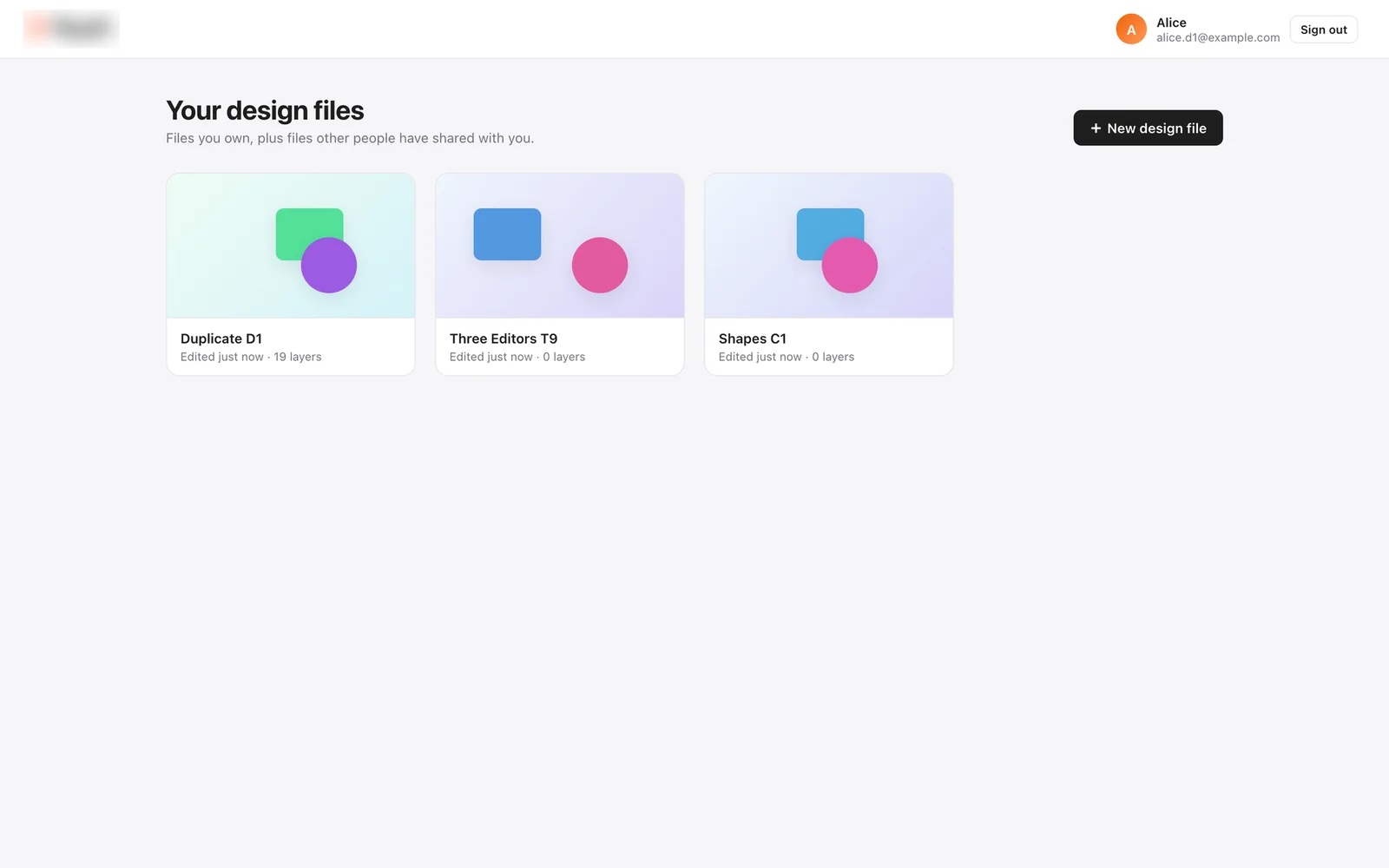} &
    \includegraphics[width=0.48\textwidth]{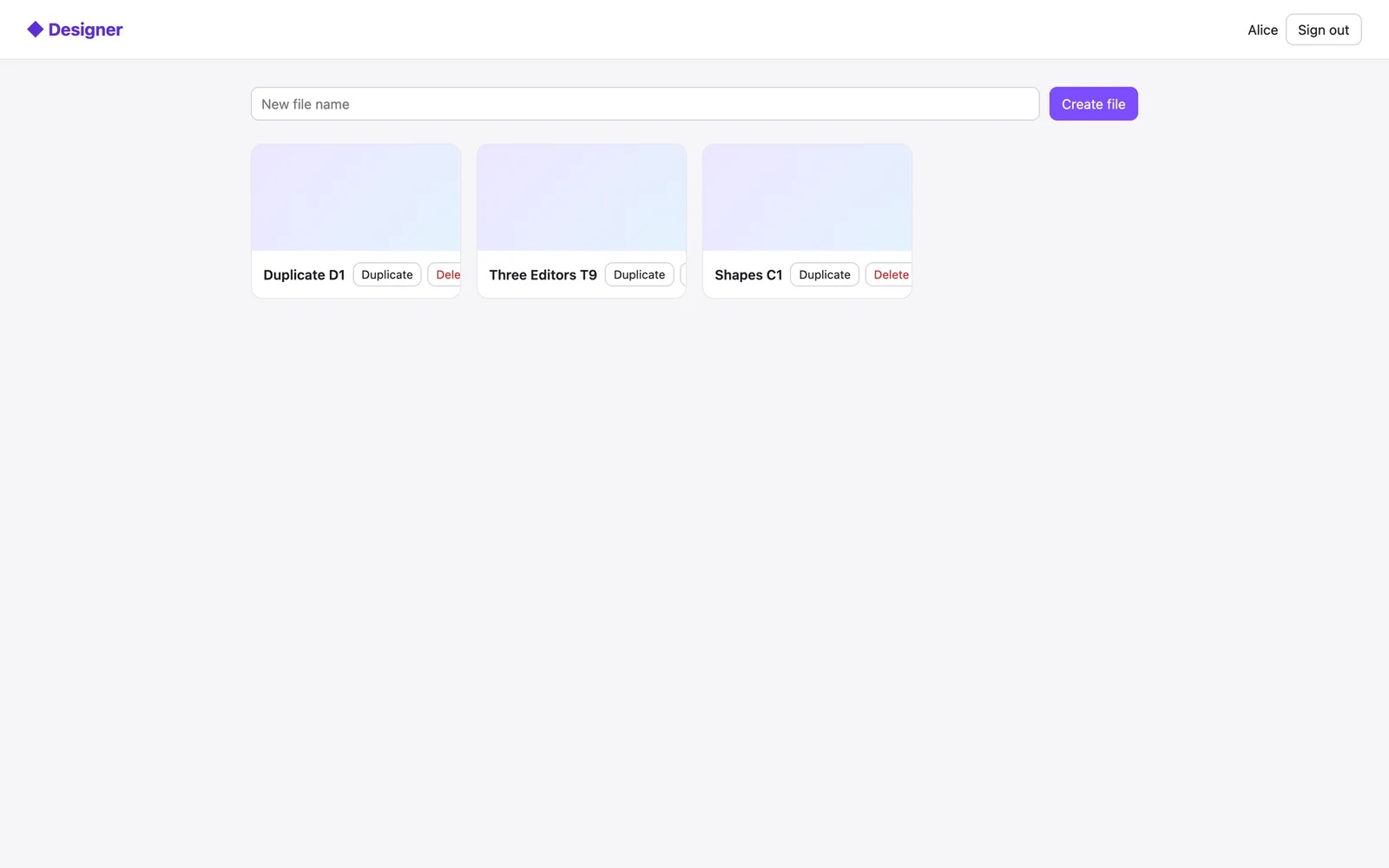} \\
    \small (c) Opus 5.5, build 1: fails 5 plans & \small (d) Sonnet 5.5, build 2: fails 8 plans \\
  \end{tabular}
  \caption{The file browser of four of our reference builds of the product modeled on Figma, each holding the same seeded files.
    Plans grade function, not visual design.
    Logos modeled on the real products are blurred.}
  \label{fig:shots}
\end{figure*}

\begin{figure*}[h]
  \centering
  \begin{tabular}{@{}cc@{}}
    \includegraphics[width=0.48\textwidth]{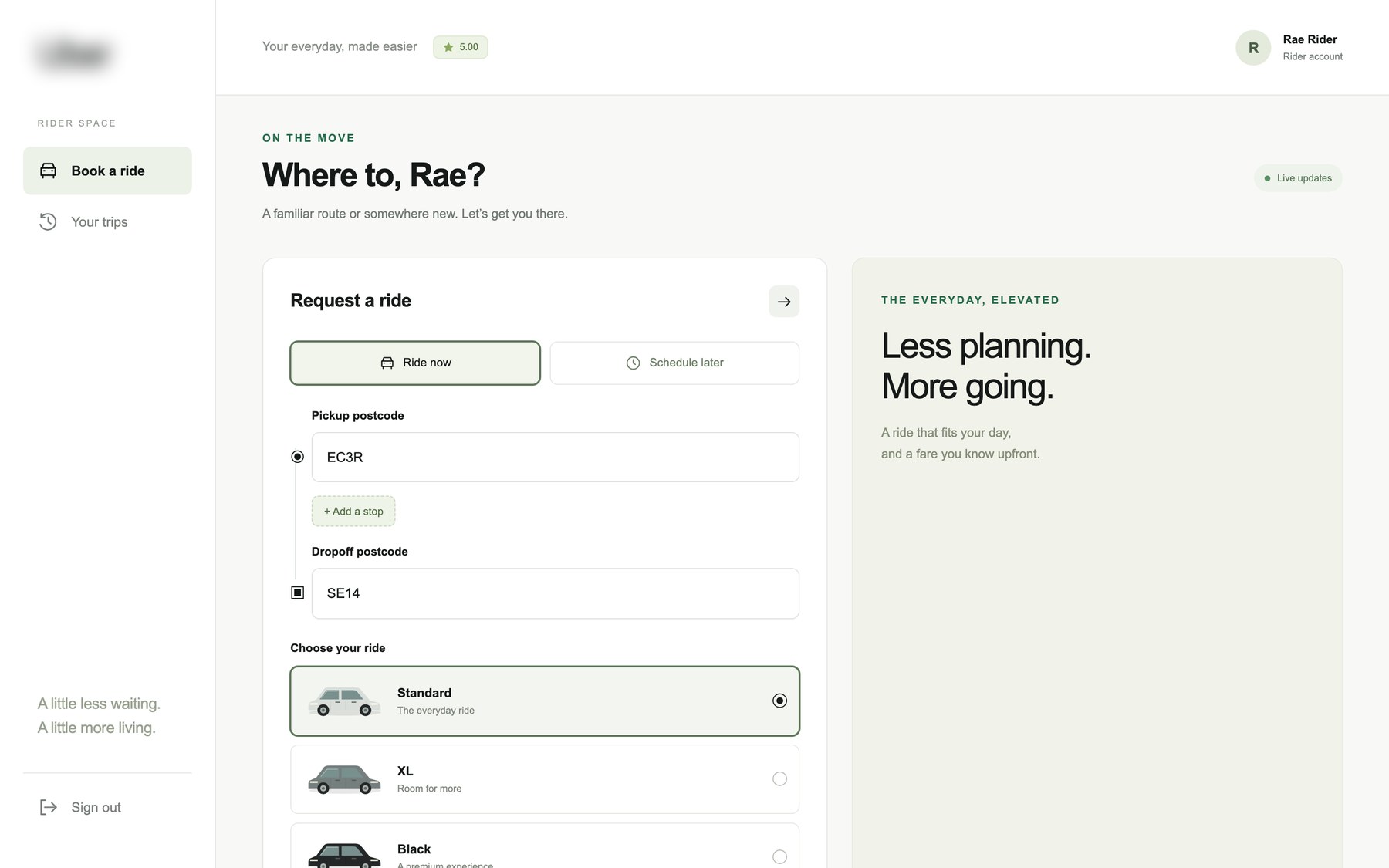} &
    \includegraphics[width=0.48\textwidth]{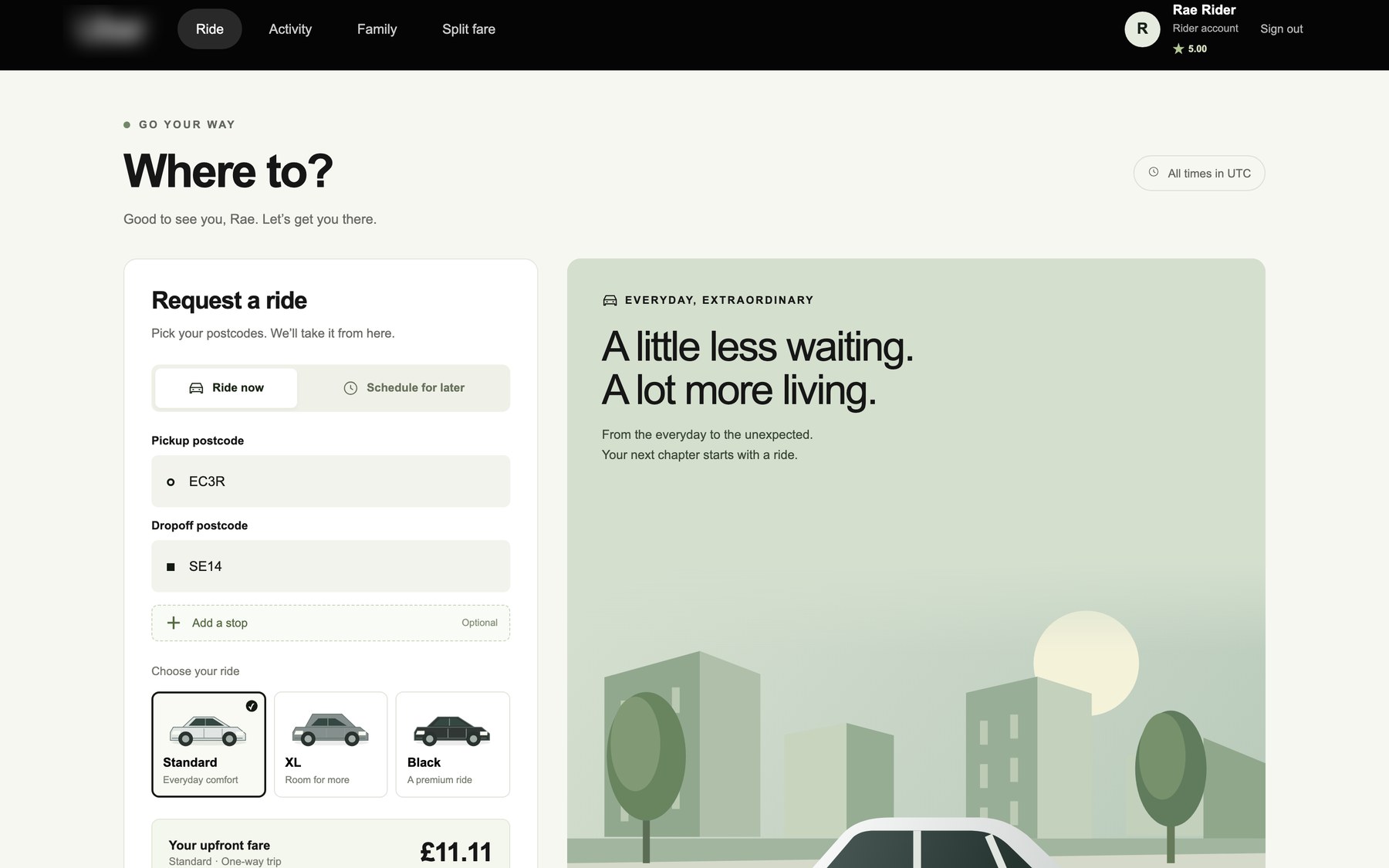} \\
    \small (a) GPT-6.1 Sol, build 1: passes every plan & \small (b) GPT-6 Astra, build 2: passes every plan \\[4pt]
    \includegraphics[width=0.48\textwidth]{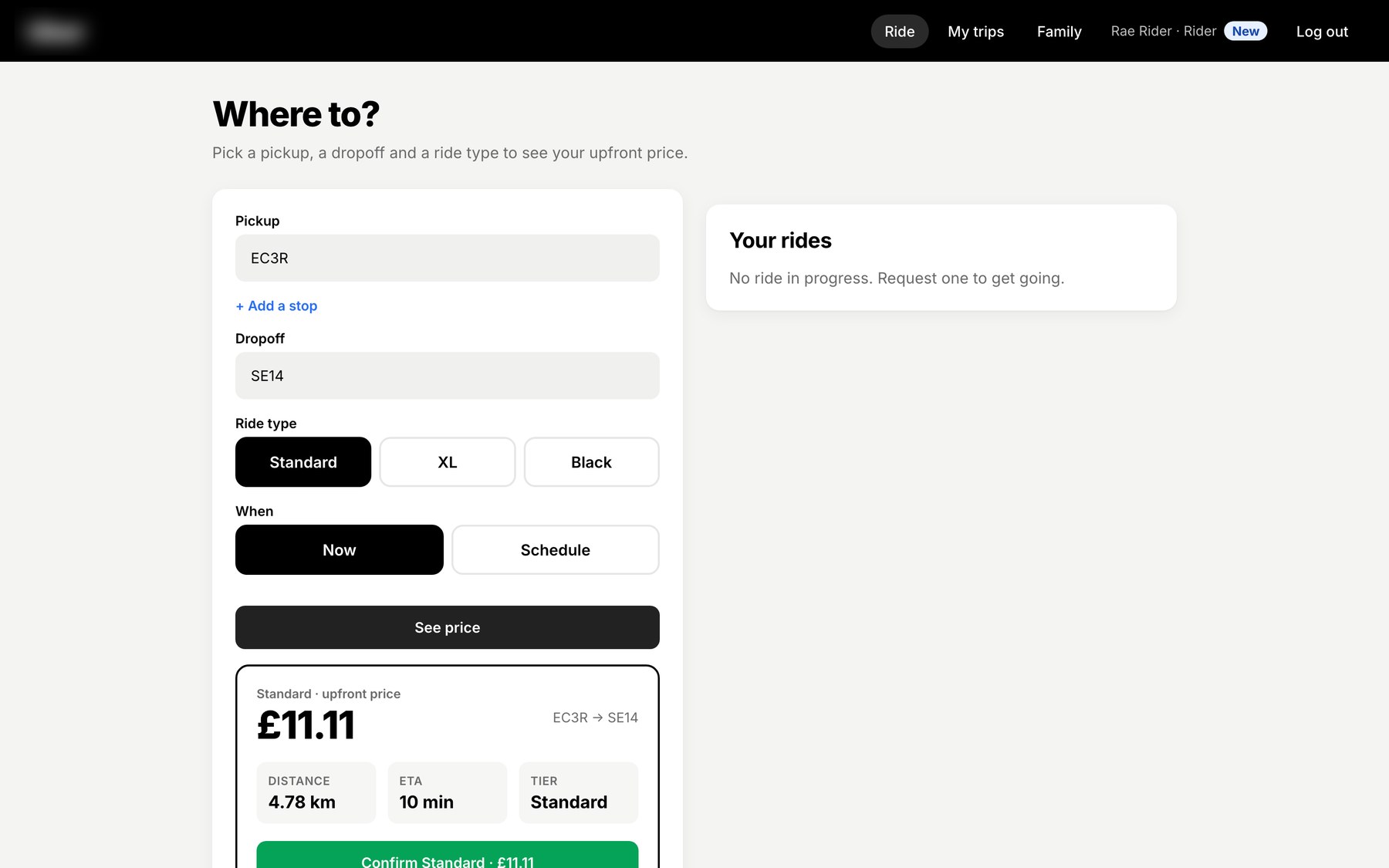} &
    \includegraphics[width=0.48\textwidth]{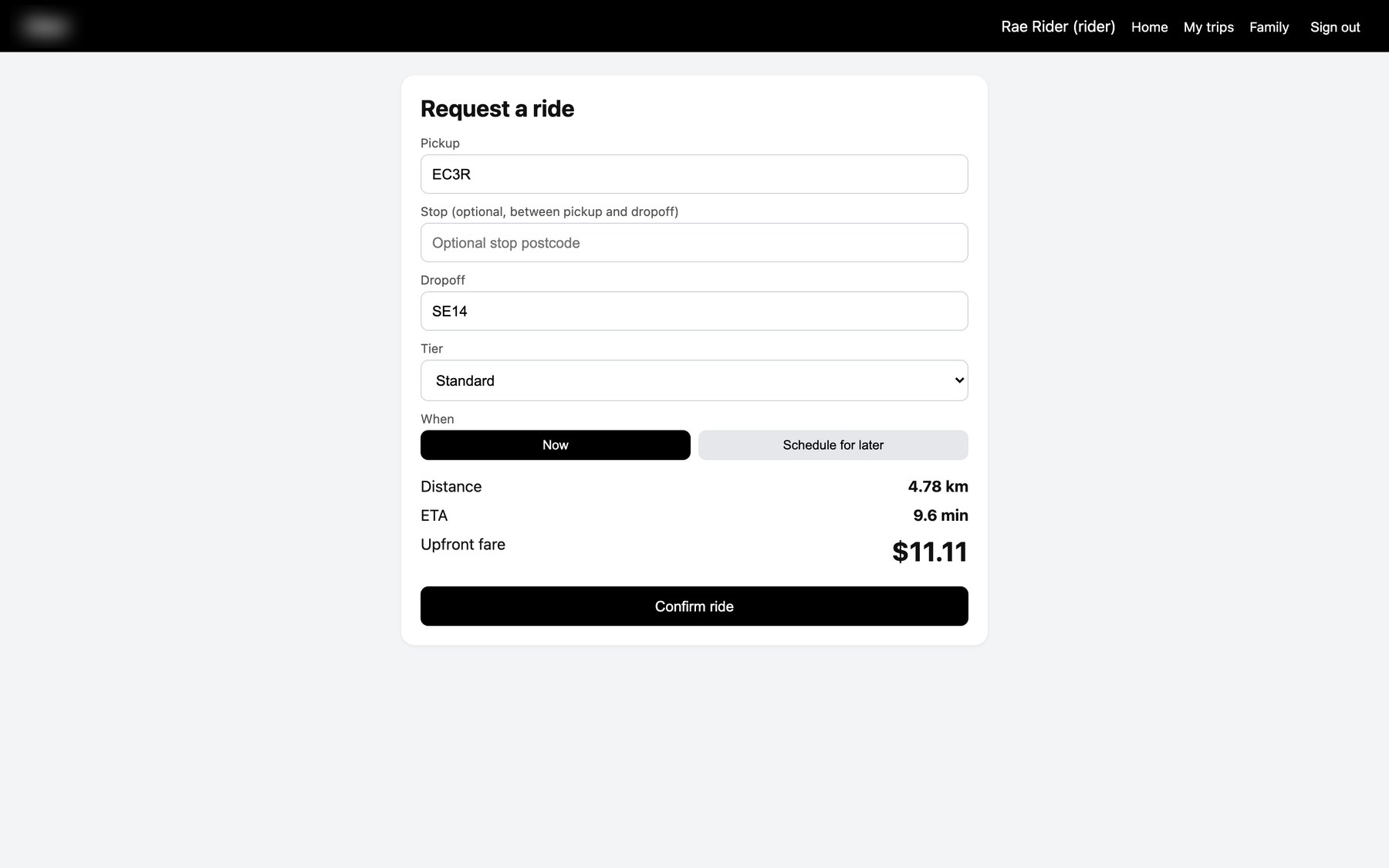} \\
    \small (c) Opus 5.5, build 1: fails 2 plans & \small (d) Sonnet 5.5, build 1: fails 2 plans \\
  \end{tabular}
  \caption{The booking page of four of our reference builds of the product modeled on Uber, with the same trip entered.
    Sonnet shows the fare in dollars, although the plans expect pounds.
    Wordmarks the builds drew are blurred.}
  \label{fig:shots-ride}
\end{figure*}

Figure~\ref{fig:shots} shows the same page of four reference builds of the product modeled on Figma, the visual comparison discussed in Section~\ref{sec:results}, and Figure~\ref{fig:shots-ride} shows the booking page of four builds of the product modeled on Uber.

\section{Additional Results}
\label{app:products}

Table~\ref{tab:products} gives the public products one by one, and Table~\ref{tab:split} compares scores on the eight public and nine held-out products.
Among the GPT releases of Table~\ref{tab:versions}, which were run on the public products, builds that do not start on an empty database become rare in later versions: 8 of 32 for GPT-5.6 Luna and 5 for GPT-5.6 Terra, against at most 1 for each later version we ran.

\begin{table*}[t]
  \centering
  \small
  \setlength{\tabcolsep}{3.5pt}
  \begin{tabular}{@{}lrrrrrrrrr@{}}
    \hline
    Model & Amazon & Asana & Discord & Figma & GitHub & Google Docs & Jira & Uber & Public 8 \\
    \hline
    \multicolumn{10}{@{}l}{\emph{pass@1: builds in which every plan passes, of 4}} \\
    GPT-6.1 Sol         & 2/4 & 4/4 & 2/4 & 0/4 & 0/4 & 1/4 & 4/4 & 4/4 & 53.1 $\pm$ 12.4 \\
    GPT-6 Astra         & 4/4 & 4/4 & 1/4 & 0/4 & 1/4 & 1/4 & 3/4 & 3/4 & 53.1 $\pm$ 14.4 \\
    Opus 5.5            & 0/4 & 2/4 & 2/4 & 0/4 & 2/4 & 0/4 & 4/4 & 2/4 & 37.5 $\pm$ 14.9 \\
    Sonnet 5.5          & 1/4 & 0/4 & 1/4 & 0/4 & 0/4 & 0/4 & 1/4 & 1/4 & 12.5 (3.5--29.0) \\
    Grok 4.7            & 0/4 & 0/4 & 0/4 & 0/4 & 0/4 & 0/4 & 1/4 & 0/4 & 3.1 (0.1--16.2) \\
    DeepSeek V4.1 Flash & 0/4 & 0/4 & 0/4 & 0/4 & 0/4 & 0/4 & 0/4 & 0/4 & 0.0 (0--10.9) \\
    GLM 5.3             & 0/4 & 0/4 & 0/4 & 0/4 & 0/4 & 0/4 & 0/4 & 0/4 & 0.0 (0--10.9) \\
    Kimi K3             & 0/4 & 0/4 & 0/4 & 0/4 & 0/4 & 0/4 & 0/4 & 0/4 & 0.0 (0--10.9) \\
    Gemini 3.8 Flash    & 0/4 & 0/4 & 0/4 & 0/4 & 0/4 & 0/4 & 0/4 & 0/4 & 0.0 (0--10.9) \\
    \hline
    \multicolumn{10}{@{}l}{\emph{Plan pass rate (\%), mean over 4 builds}} \\
    GPT-6.1 Sol         & 90.7 & 100.0 & 92.6 & 92.2 & 97.0 & 86.7 & 100.0 & 100.0 & 94.9 $\pm$ 2.5 \\
    GPT-6 Astra         & 100.0 & 100.0 & 94.9 & 92.2 & 98.2 & 94.2 & 99.2 & 99.1 & 97.2 $\pm$ 1.1 \\
    Opus 5.5            & 91.7 & 90.8 & 97.8 & 88.8 & 73.8 & 91.7 & 100.0 & 98.3 & 91.6 $\pm$ 6.6 \\
    Sonnet 5.5          & 92.6 & 84.5 & 94.9 & 47.4 & 89.6 & 80.8 & 68.8 & 69.8 & 78.7 $\pm$ 9.8 \\
    Grok 4.7            & 25.9 & 19.2 & 50.7 & 50.9 & 76.8 & 12.5 & 44.5 & 48.3 & 41.1 $\pm$ 13.7 \\
    DeepSeek V4.1 Flash & 55.6 & 77.4 & 85.3 & 62.1 & 69.5 & 60.0 & 78.9 & 87.7 & 72.1 $\pm$ 5.9 \\
    GLM 5.3             & 73.1 & 88.3 & 79.4 & 67.2 & 70.1 & 31.7 & 76.6 & 20.7 & 63.4 $\pm$ 6.8 \\
    Kimi K3             & 21.3 & 52.5 & 50.7 & 47.4 & 76.7 & 30.4 & 86.7 & 64.7 & 53.8 $\pm$ 12.7 \\
    Gemini 3.8 Flash    & 19.4 & 7.5 & 13.2 & 16.4 & 25.6 & 23.3 & 33.6 & 86.2 & 28.2 $\pm$ 7.7 \\
    \hline
  \end{tabular}
  \caption{Results of the independent runs on each public product, 4 builds per product (32 builds per model).
    The last column is the public total in percent with its 95\% interval; near 0\% we give the exact binomial range.
    Held-out products are reported only in total (Table~\ref{tab:split}).}
  \label{tab:products}
\end{table*}

\begin{table}[tb]
  \centering
  \small
  \setlength{\tabcolsep}{3pt}
  \begin{tabular}{@{}lrrrr@{}}
    \hline
    & \multicolumn{2}{c}{pass@1} & \multicolumn{2}{c}{Plan pass rate} \\
    Model & Public & Held out & Public & Held out \\
    \hline
    GPT-6.1 Sol         & 53.1 & 75.0 & 94.9 & 98.7 \\
    GPT-6 Astra         & 53.1 & 66.7 & 97.2 & 98.7 \\
    Opus 5.5            & 37.5 & 58.3 & 91.6 & 95.7 \\
    Sonnet 5.5          & 12.5 & 25.0 & 78.7 & 79.6 \\
    Grok 4.7            & 3.1  & 0.0  & 41.1 & 63.3 \\
    DeepSeek V4.1 Flash & 0.0  & 0.0  & 72.1 & 82.2 \\
    GLM 5.3             & 0.0  & 0.0  & 63.4 & 68.4 \\
    Kimi K3             & 0.0  & 0.0  & 53.8 & 59.4 \\
    Gemini 3.8 Flash    & 0.0  & 0.0  & 28.2 & 46.9 \\
    \hline
  \end{tabular}
  \caption{Scores of the independent runs on the 8 public products (32 builds per model) and the 9 held-out products (36 builds).
    Table~\ref{tab:products} gives the public products one by one.}
  \label{tab:split}
\end{table}

\section{ViBench 1.0 Rerun}
\label{app:rerun}

We re-ran the Vibe-on-Vibe tasks of ViBench~\citep{zhong2026vibench} with four of the models in Table~\ref{tab:main}: each model first builds the MVPs of 15 apps, then adds each of 45 features to its own MVP in a fresh conversation, once per feature, and the feature builds are graded on 152 plans.
We kept ViBench's harness, seeding agent, grader and single grade per plan; builders ran at medium reasoning effort, as in Table~\ref{tab:main}, whereas ViBench's release used each model's highest.
Table~\ref{tab:rerun} gives pass@1; 11 of the 608 plans failed seeding, 9 of them on Claude builds of one app, and count as failed, as in ViBench.

\begin{table}[ht]
  \centering
  \small
  \begin{tabular}{@{}lr@{}}
    \hline
    Model & pass@1 \\
    \hline
    GPT-6.1 Sol     & 88.9 \\
    GPT-6 Astra     & 82.2 \\
    Opus 5.5        & 82.2 \\
    Sonnet 5.5      & 60.0 \\
    \hline
    Opus 4.6$^\ast$ & 51.1 \\
    \hline
  \end{tabular}
  \caption{Our ViBench 1.0 rerun on its Vibe-on-Vibe tasks: pass@1 in percent, one build per feature (45 per model).
    $^\ast$The best model at ViBench's release in its standard harness, as reported by \citet{zhong2026vibench}.}
  \label{tab:rerun}
\end{table}

\end{document}